\documentclass[10pt,twocolumn,letterpaper]{article}
\usepackage[page]{appendix}
\usepackage{tocbibind}
\usepackage{authblk}
\usepackage[utf8]{inputenc}
\usepackage[mathletters]{ucs}
\usepackage[T1]{fontenc}
\usepackage{amsmath,amssymb,amsfonts}
\usepackage{graphicx}
\usepackage{booktabs}
\graphicspath{{images/}}
\usepackage{xcolor}
\usepackage{tabularx}
\usepackage[top=1in, bottom=1in, left=1in, right=1in]{geometry}
\usepackage{abstract}
\usepackage{floatrow}
\usepackage{float}
\usepackage{array}
\usepackage{multirow}
\usepackage{placeins}
\restylefloat{table}
\usepackage[numbers,sort&compress]{natbib}
\setcitestyle{sort}
\usepackage{hyperref}
\usepackage[capitalise]{cleveref}
\makeatletter
\let\@fnsymbol\@alph
\makeatother

\title{Public Domain 12M: A Highly Aesthetic Image-Text Dataset with Novel Governance Mechanisms}

\author{Jordan Meyer}
\author{Nick Padgett}
\author{Cullen Miller}
\author{Laura Exline}
\affil{Spawning}

\begin{document}

\twocolumn[
  \begin{@twocolumnfalse}
    \maketitle
    \begin{abstract}
    \noindent
We present Public Domain 12M (PD12M), a dataset of 12.4 million high-quality public domain and CC0-licensed images with synthetic captions, designed for training text-to-image models. PD12M is the largest public domain image-text dataset to date, with sufficient size to train foundation models while minimizing copyright concerns. Through the Source.Plus platform, we also introduce novel, community-driven dataset governance mechanisms that reduce harm and support reproducibility over time.  
\end{abstract}
\end{@twocolumnfalse}
\vspace{1cm}
]

\section{Introduction}

Advancements in computer vision and natural language processing have fueled demand for ever larger image-text datasets \cite{sharma2018conceptual, thomee2016yfcc100m, schuhmann2021laion400m, schuhmann2022laion, schuhmann2023datacomp, alrashdi2023coyo} to train increasingly sophisticated models \cite{radford2021learning, ramesh2022hierarchical, rombach2022high}. To meet this demand, large-scale datasets typically comprise URLs identified by web crawlers and require each model trainer to re-download the images from across the web. This practice has been critiqued for prioritizing scale over responsibility \cite{prabhu2023hatescaling} and has prompted a range of concerns regarding copyright infringement \cite{longpre2023largescale, szkalej2024mapping}, consent \cite{prabhu2020large, longpre2023consent}, inappropriate content \cite{laion2023maintenance, birhane2021multimodal}, personally identifiable information (PII) \cite{paullada2021data}, and the perpetuation of harmful biases \cite{wang2020revise, crawford2021excavating}. Moreover, once released, these datasets degrade over time \cite{carlini2023poisoning} and are ill-equipped to remediate their contents \cite{mitchell2023data}.

Responding to these concerns, we introduce a new public domain dataset and a novel platform for dataset governance that incorporate current recommendations for responsible dataset creation and maintenance.

Public Domain 12M (PD12M) is the largest image-text dataset sourced from only materials labeled with a \href{https://creativecommons.org/public-domain/pdm/}{Public Domain Mark} or Creative Commons Zero license (\href{https://creativecommons.org/public-domain/cc0/}{CC0}). At 12.4 million image-caption pairs, PD12M and its 3.3 million item subset, Public Domain 3M (PD3M), match the size of the Conceptual Captions datasets (CC12M and CC3M) \cite{sharma2018conceptual}, enabling commercial implementations of AI models trained using those and similarly sized datasets \cite{gao2024lumina, chen2023pixart, mitsua2024diffusion, gu2023matryoshka}. PD12M and PD3M are released under a Community Data License Agreement (\href{https://cdla.dev/permissive-2-0/}{CDLA-Permissive-2.0}). 

In this paper, we describe our image sourcing and curation process, which was designed to maximize the aesthetic quality of the dataset contents while minimizing copyright concerns. Because the problems of web-scale datasets go well beyond copyright \cite{prabhu2020large, birhane2021multimodal, longpre2023consent, crawford2021excavating}, we also outline recommendations for dataset governance and explain how we implement those recommendations with the \href{https://source.plus}{Source.Plus} platform.

\begin{table*}[!ht]
\centering
\small
\setlength{\tabcolsep}{4pt}
\begin{tabular}{l|c|c|c|c}
\toprule
\textbf{Dataset} & \textbf{Scale} & \textbf{Licenses} & \textbf{Caption} & \textbf{Governance} \\
\midrule
Megalith-10m \cite{megalith2024} & 10M & PD, CC0, "No known copyright" & Synthetic & None \\
\hline
CC12M \cite{sharma2018conceptual} & 12.4M & Copyrighted \& opted-out images & Alt-text & None \\
\hline
\textbf{PD12M} & \textbf{12.4M} & \textbf{PD/CC0 only} & \textbf{Synthetic} & \textbf{Community} \\
\hline
CommonCatalog \cite{gokaslan2023commoncanvas} & 70M & Mixed CC licenses & Synthetic & Limited \\
\hline
YFCC100M \cite{thomee2016yfcc100m} & 100M & Mixed CC licenses & Description & None \\
\hline
LAION-400M \cite{schuhmann2021laion400m} & 400M & Copyrighted \& opted-out images & Alt-text & Limited \\
\hline
COYO-700M \cite{alrashdi2023coyo} & 700M & Copyrighted \& opted-out images & Alt-text & None \\
\hline
LAION-5B \cite{schuhmann2022laion} & 5.85B & Copyrighted \& opted-out images & Alt-text & Limited \\
\hline
DataComp-12B \cite{schuhmann2023datacomp} & 12.8B & Copyrighted \& opted-out images & Alt-text & Limited \\
\bottomrule
\end{tabular}
\caption{Comparison of major image-text datasets highlighting key differences. While several datasets attempt to address licensing with CC sources or address scale with web scraping, PD12M uniquely combines clear licensing and formal governance mechanisms for ongoing dataset maintenance. We were unable to find governance documentation for four datasets; datasets with "Limited" governance have narrowly scoped takedown protocols, e.g., to remove CSAM.}
\label{tab:dataset_comparison}
\end{table*}

\section{Background}

Datasets like CommonCatalog \cite{gokaslan2023commoncanvas} and Megalith-10M \cite{megalith2024} have attempted to address copyright concerns by including only Creative Commons (CC) licenses. However, the ShareAlike and Attribution requirements of many CC licenses raise unresolved legal questions regarding whether model outputs can satisfy attribution requirements and whether derivative works created by AI models must carry the same license terms \cite{longpre2023largescale, szkalej2024mapping}. 

Beyond copyright concerns, proponents of responsible AI have proposed methods for identifying and mitigating dataset biases \cite{wang2020revise, mitchell2023data} and measuring distribution shifts and data quality \cite{rabanser2019failing, lipton2018detecting}. Additional work has outlined documentation standards that seek to increase dataset transparency and allow researchers to make more educated decisions about possible limitations and biases that a specific dataset might introduce \cite{gebru2021datasheets, mitchell2019model, bender2018data, holland2018dataset}. Researchers have also emphasized the need for sustainable dataset maintenance \cite{wilkinson2016fair}, refinement \cite{bender2018data}, and quality preservation \cite{mitchell2019model}. 

Additionally, the emerging field of public AI infrastructure offers new frameworks to better protect AI resources as public goods \cite{eaves2024digital, chan2023reclaiming, tarkowski2022public, keller2022public}. The BigScience ROOTS corpus \cite{bigscience2023roots} and Mozilla Common Voice \cite{ardila2020common} demonstrate how public datasets can be developed through open collaboration while adhering to FAIR data principles \cite{wilkinson2016fair}. These projects align with emerging frameworks for Open Source AI \cite{osi2024openaiai} by prioritizing transparency \cite{hutson2018artificial} and democratic participation \cite{gilman2023democratizing, seger2023democratising} in their development processes.

These recommendations guided our choices throughout the image collection and curation process for PD12M, and our platform for dataset governance incorporates mechanisms that engage community participation and promote Public AI.

\section{Building PD12M}

We include a link to an up-to-date dataset summary in Appendix \ref{app:summary}. We also completed a datasheet \cite{gebru2021datasheets} and included it alongside the dataset itself on Hugging Face.

\subsection{Image Collection}

\textbf{GLAM Metadata:} We sourced 23.1M images directly from galleries, libraries, archives, and museums (GLAM), as well as from aggregators of their content. GLAM institutions provide additional layers of quality, safety, and licensing review \cite{tarkowski2022public} and more comprehensive information about an item's origins and authenticity \cite{holland2018dataset, mitchell2023data} when compared to Common Crawl \cite{luccioni2023undesirable}.  The OpenGLAM Survey \cite{mccarthy2024openglam}, which tracks open access policies across cultural heritage institutions worldwide, provides the most thorough guidance for locating these organizations. A comprehensive list of all contributing institutions and the volume of their contributions, is linked in Appendix~\ref{app:summary}. 

We created custom parsers for each source to extract provenance, metadata, and licensing information for every image we collected \cite{longpre2023largescale, bender2018data, mitchell2018web, jo2020lessons} from OpenGLAM sources. 

We parsed metadata from static files when available, and when traversing APIs, we used self-imposed rate limiting. These measures were taken to minimize the strain on the hosting institutions' servers \cite{holscher2024crawlers}.

Prior to cloning any images, we filtered the extracted metadata to include only images explicitly marked as public domain or CC0 by the source. See Appendix~\ref{app:disclaimer}. 

\textbf{Wikimedia Metadata:} We sourced 11.3M images from Wikimedia Commons \cite{wikimedia2023}, which contains a mix of user uploads and institutional datasets. While the Wikimedia Commons community rigorously maintains their media collections, we took additional steps to pre-filter undesirable content based on the metadata.

We first limited our metadata collection to the \href{https://commons.wikimedia.org/wiki/Category:CC-PD-Mark}{CC-PD-Mark} and \href{https://commons.wikimedia.org/wiki/Category:CC-Zero}{CC-Zero} categories only. Wikimedia users can tag items with more granular information. We manually reviewed the user tags within these categories and filtered out any images with tags that implied a license outside of the public domain. Additionally, we filtered out images tagged as AI-generated or as having other usage restrictions beyond copyright. A full list of the excluded tags can be found in Appendix~\ref{app:summary}.

We also implemented a two-week delay between parsing the metadata and ingesting images from Wikimedia Commons. This delay provided time for the Wikimedia community's robust moderation process to vet newer content before our final ingestion. 

\textbf{iNaturalist Metadata:} We sourced 3.2M images from iNaturalist, which were contributed by individual members of its community. Our initial review of the CC0 subset of iNaturalist found no problematic images. Nevertheless, we also implemented a two-week delay between parsing the iNaturalist metadata and ingesting  images to provide time for newer submissions to be vetted by its community moderation process.

\textbf{Validating Rights Reservations:} The European Union provides a copyright exception for commercial Text and Data Mining, so long as rights reservations (opt-outs) are respected \cite{eu2019copyright}. We cross-referenced the image URLs from the above sources against the $\sim$2B URLs covered by Spawning's Do Not Train Registry (DNTR) \cite{spawning2024dnt} and found that none were opted out of AI training as of October 2024. In contrast, analysis by Hugging Face found that 25.4\% of URLs in CC3M were opted out when checked against the DNTR \cite{huggingface2024cc}. 

\textbf{Image Downloading:}
We collected 38M image URLs and their metadata from the sources above. Our final ingestion process included downloading and storing the identified images. We used self-imposed rate limiting to avoid overburdening the hosts, spreading image downloads across two months. The complete collection of these images and their metadata is available for review on Source.Plus.

\subsection{Image Curation}

We used automated and manual  filtering to ensure the overall quality and safety of the dataset. Our curation process was primarily performed through Source.Plus. We filtered the 38M downloaded images described in the previous section down to 12.4M for PD12M and to 3.3M for the PD3M subset.

\textbf{Semantic Embeddings:} We first used CLIP ViT-L/14 \cite{openai2021clip} to generate embeddings, which served as inputs for downstream curation tasks, including document scan identification, NSFW filtering, and aesthetic scoring. We include these embeddings alongside the dataset.

\textbf{Document Scan Identification:} After visually inspecting the collected images, we estimated that roughly a quarter were scans of documents, such as books, letters, and newspapers, which pictured only typed or written words. For the first filtering step, we tagged 8.7M images using an internal model trained to identify document scans and removed them from some subsequent tasks in order to reduce compute costs.  

\textbf{Format and Resolution Restrictions:} We then imposed a minimum resolution threshold of 256x256 pixels to maximize the dataset's suitability for increasingly high-resolution training of text-to-image models \cite{xie2024sana, chen2024pixartsigmaweaktostrongtrainingdiffusion}.

\textbf{Content Filtering:} We used LAION's NSFW classifier \cite{schonfeld2021laion} to exclude works with a score greater than 0.5. We also manually reviewed the dataset using semantic search tools to remove instances of non-artistic photographic nudity.

Additionally, we completed a manual check of known ethnophaulisms listed in Wikipedia \cite{wikipedia2024ethnophaulisms}. For each term, we searched metadata and conducted a semantic search to remove derogatory metadata and images \cite{luccioni2023undesirable, birhane2021multimodal}. 

These steps flagged fewer than 0.05\% of the 38M total images collected, demonstrating the value of limiting our initial image collection to trusted and moderated sources.

\textbf{Deduplication:} We represented the dataset's images as nodes in a sparse graph, with links created between images when the cosine distance of their SSCD embeddings \cite{pizzi2022sscd} was <0.1 (empirically determined). Each subgraph with more than one member was treated as a group of duplicates. For each duplicate group, we selected a canonical image by choosing the item having, in order of priority: 1) a GLAM source, if available, 2)  the largest image dimensions, 3) the highest aesthetic score, 4) the largest file size, and/or 5) the most complete metadata.

\textbf{Aesthetic Scoring:}  We assigned each downloaded image an aesthetic score using a model trained on an internal dataset of human ratings. For our final curation step, to match the original size of CC12M (12.4M), we excluded images from the bottom 50\% of aesthetic scores. To match the original size of CC3M (3.3M), we excluded the bottom 90\% of images by aesthetic scores.  Randomly sampled images from select percentile ranges are available in Table~\ref{tab:aesthetic_samples} to demonstrate the model's aesthetic preferences. 

\begin{table*}[!ht]
\centering
\setlength{\tabcolsep}{3pt}
\caption{Distribution of aesthetic scores across PD12M and PD3M, with randomly sampled images from each percentile range. PD12M includes images from the 50th to 100th percentile range. PD3M includes images from the 90th to 100th percentile range.}
\begin{tabularx}{\textwidth}{@{}X|*{6}{>{\centering\arraybackslash}m{1.8cm}}@{}}
\toprule
\textbf{\footnotesize Aesthetic Range} & \textbf{\footnotesize } & \textbf{\footnotesize } & \textbf{\footnotesize } & \textbf{\footnotesize } & \textbf{\footnotesize } & \textbf{\footnotesize } \\
\midrule
{\footnotesize \raggedright \textbf{90th - 100th percentile}

3,301,972 items

27\% of PD12M | 100\% of PD3M} &
\raisebox{-.5\height}{\includegraphics[width=1.8cm,height=1.8cm,keepaspectratio]{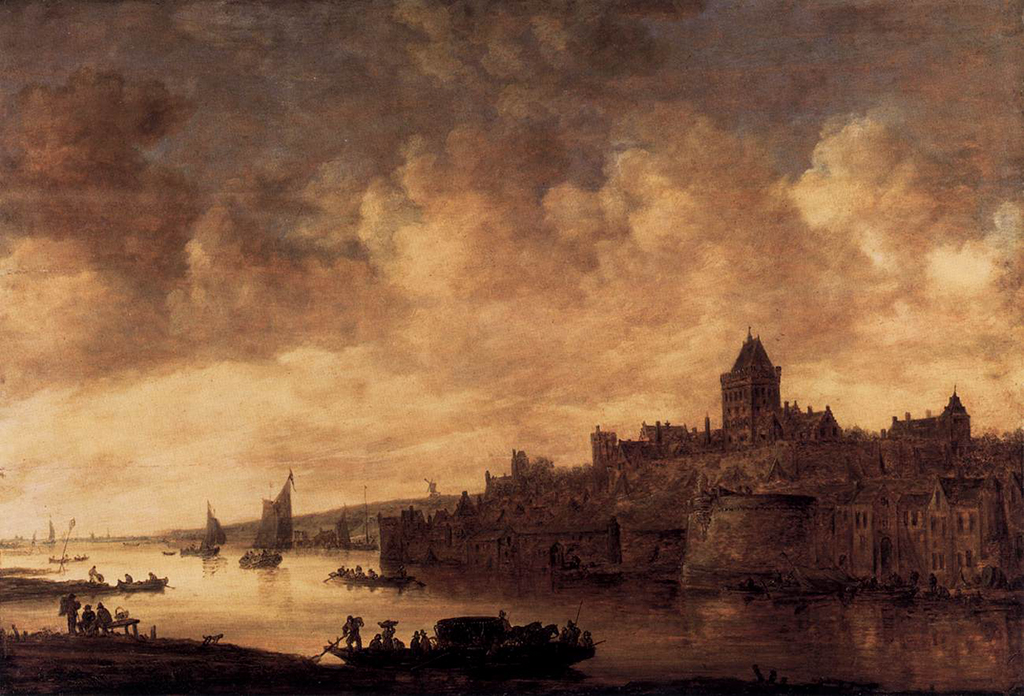}} &
\raisebox{-.5\height}{\includegraphics[width=1.8cm,height=1.8cm,keepaspectratio]{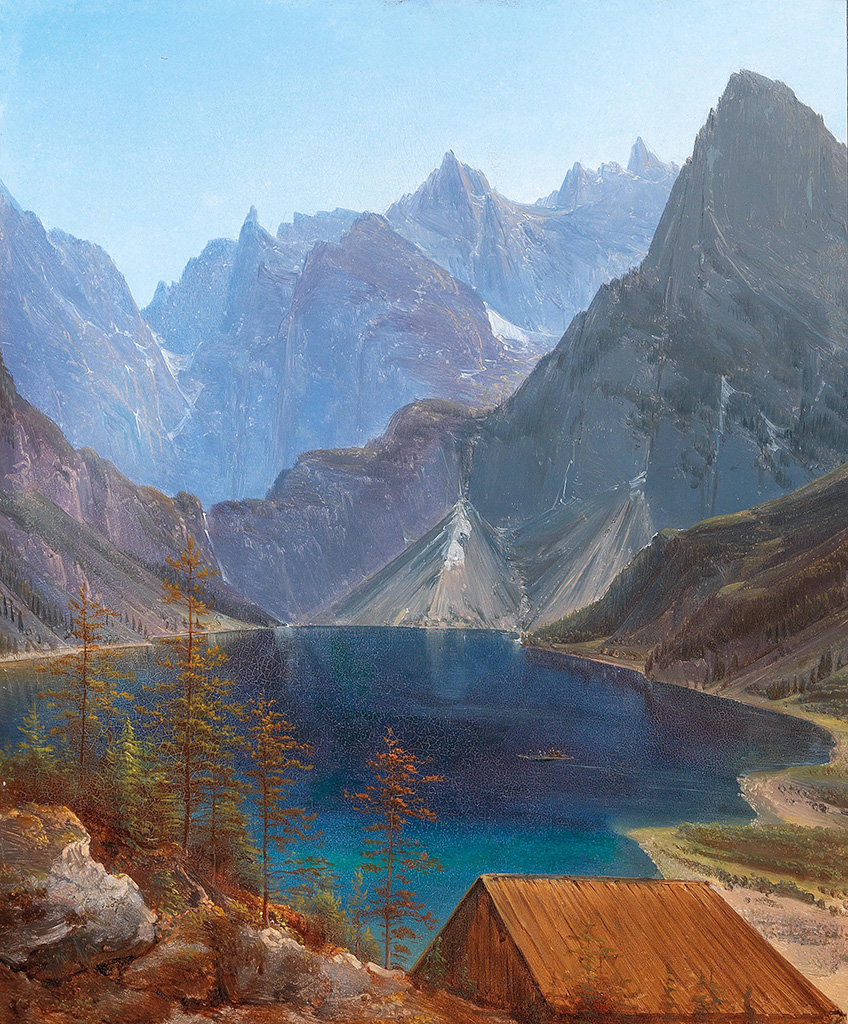}} &
\raisebox{-.5\height}{\includegraphics[width=1.8cm,height=1.8cm,keepaspectratio]{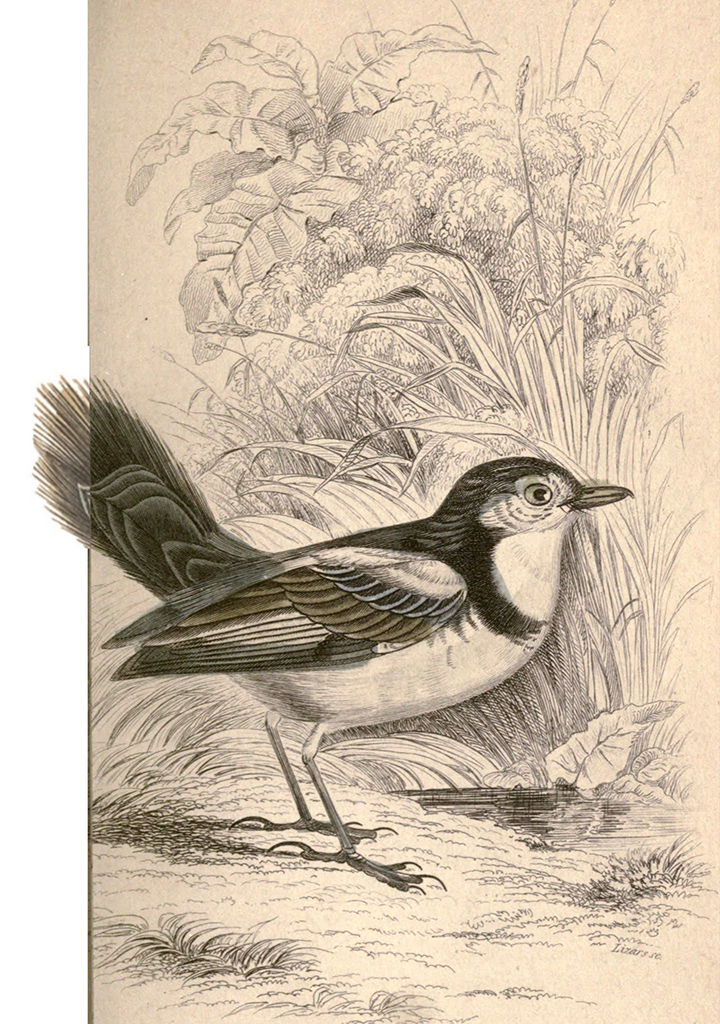}} &
\raisebox{-.5\height}{\includegraphics[width=1.8cm,height=1.8cm,keepaspectratio]{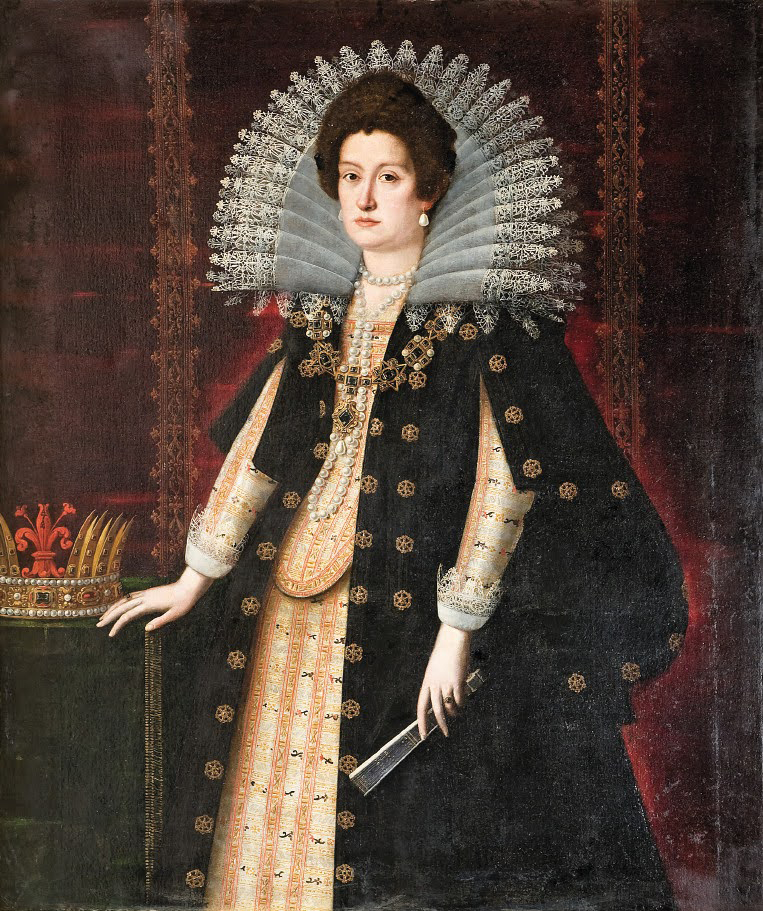}} &
\raisebox{-.5\height}{\includegraphics[width=1.8cm,height=1.8cm,keepaspectratio]{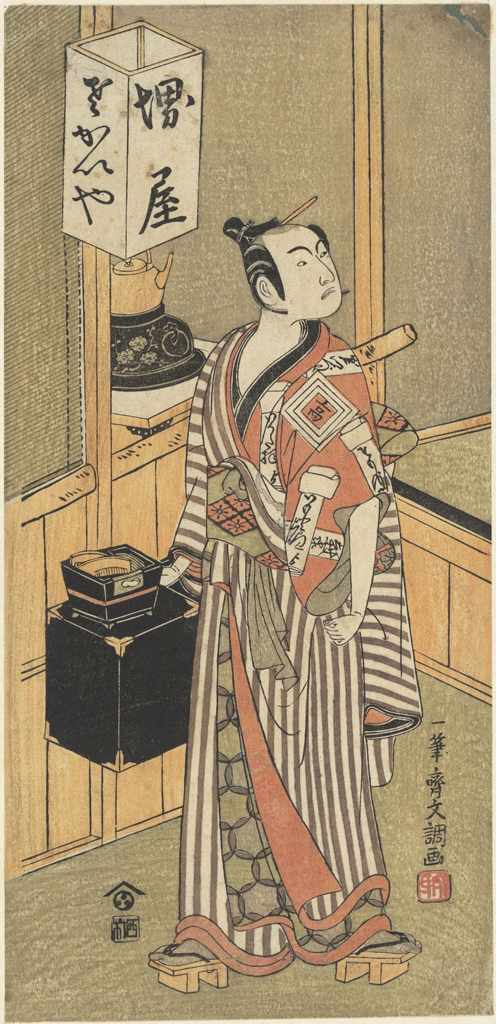}} &
\raisebox{-.5\height}{\includegraphics[width=1.8cm,height=1.8cm,keepaspectratio]{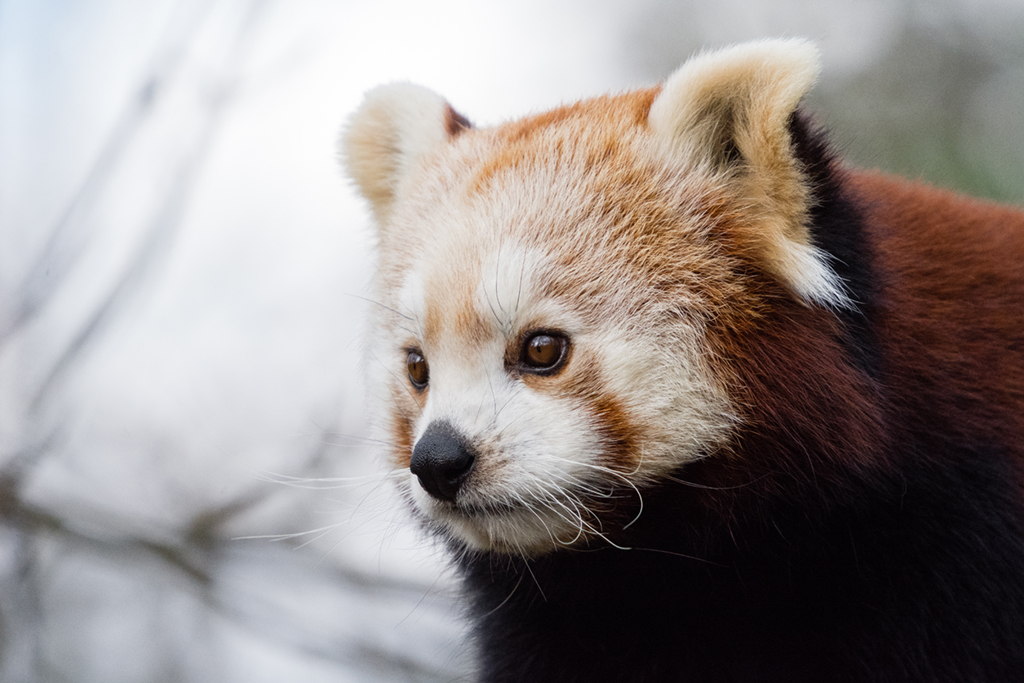}} \\
\midrule
{\footnotesize \raggedright \textbf{80th - 90th percentile}

3,060,789 items

25\% of PD12M} &
\raisebox{-.5\height}{\includegraphics[width=1.8cm,height=1.8cm,keepaspectratio]{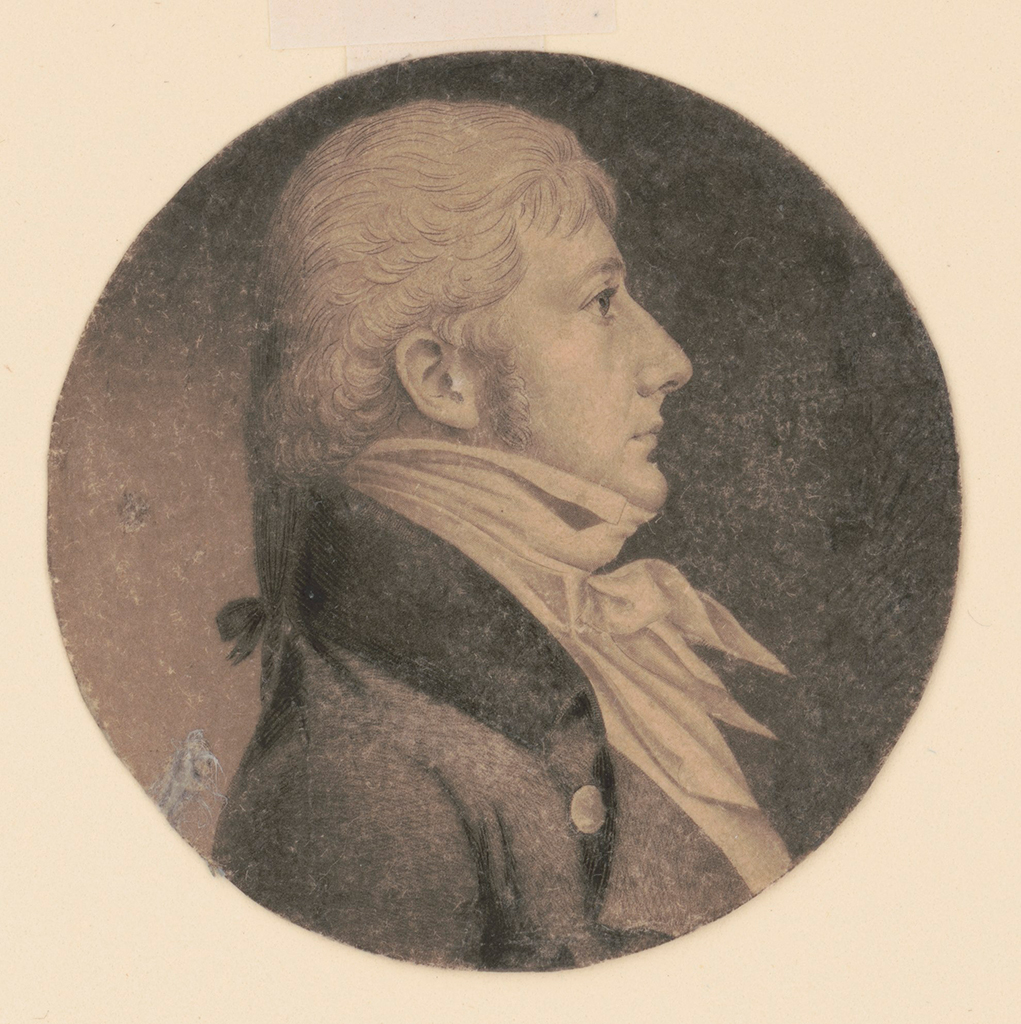}} &
\raisebox{-.5\height}{\includegraphics[width=1.8cm,height=1.8cm,keepaspectratio]{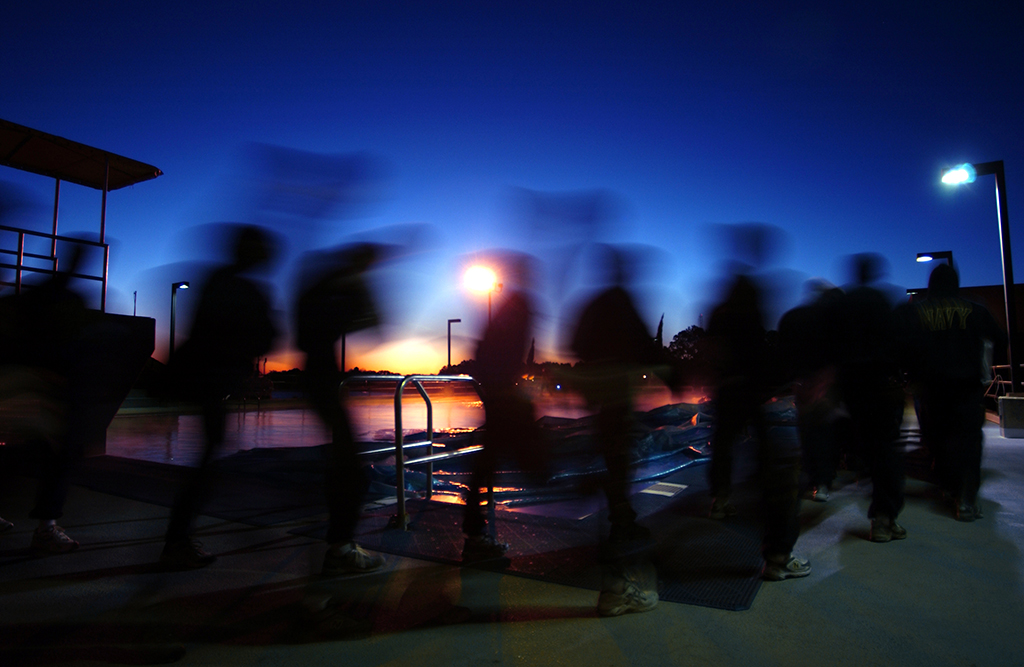}} &
\raisebox{-.5\height}{\includegraphics[width=1.8cm,height=1.8cm,keepaspectratio]{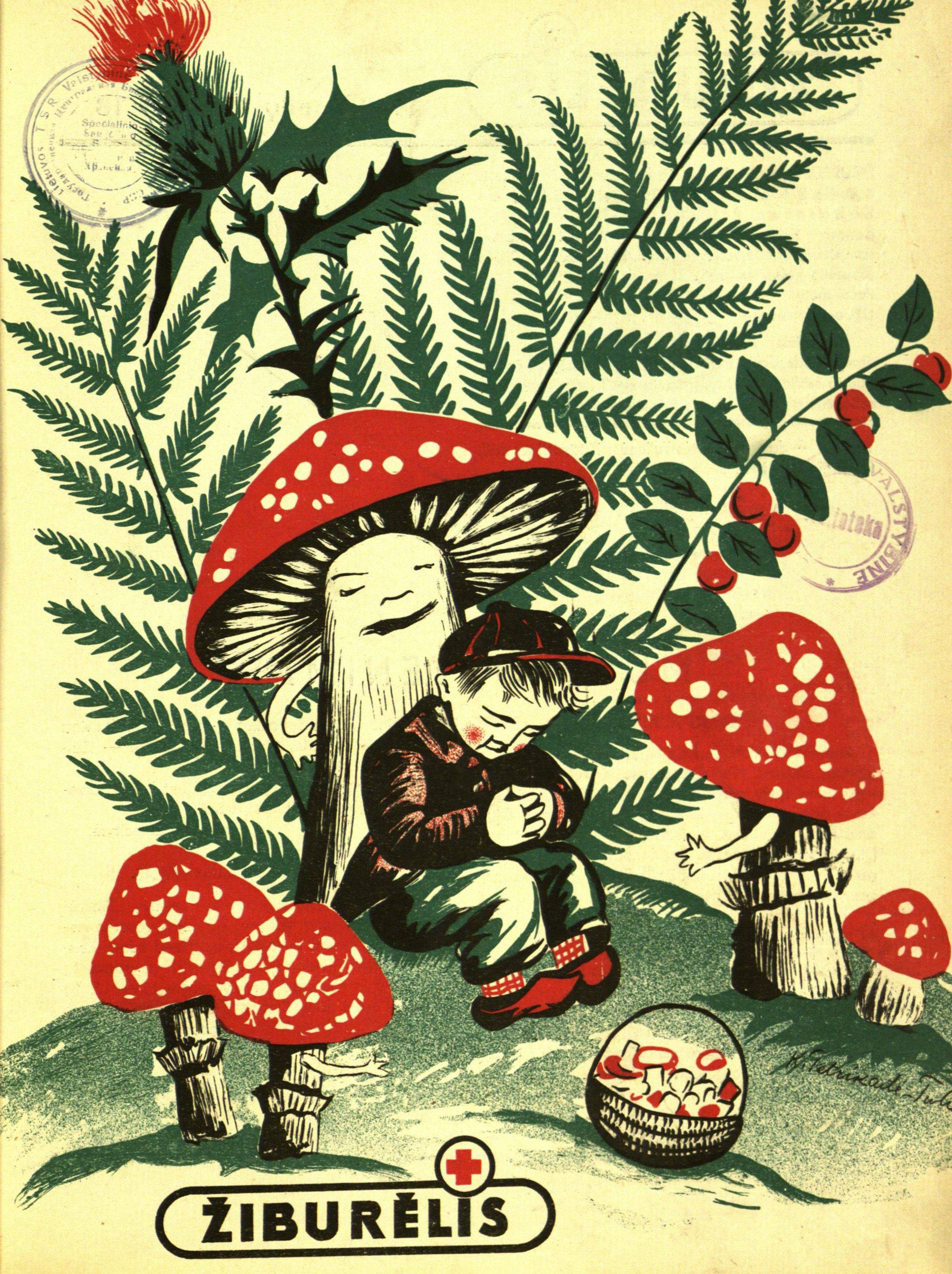}} &
\raisebox{-.5\height}{\includegraphics[width=1.8cm,height=1.8cm,keepaspectratio]{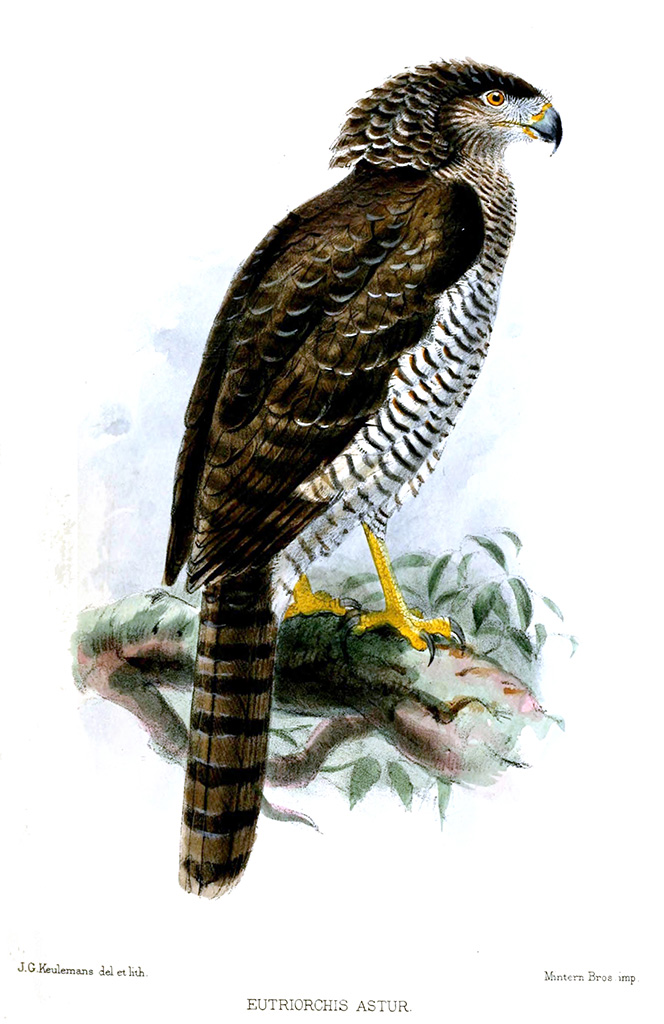}} &
\raisebox{-.5\height}{\includegraphics[width=1.8cm,height=1.8cm,keepaspectratio]{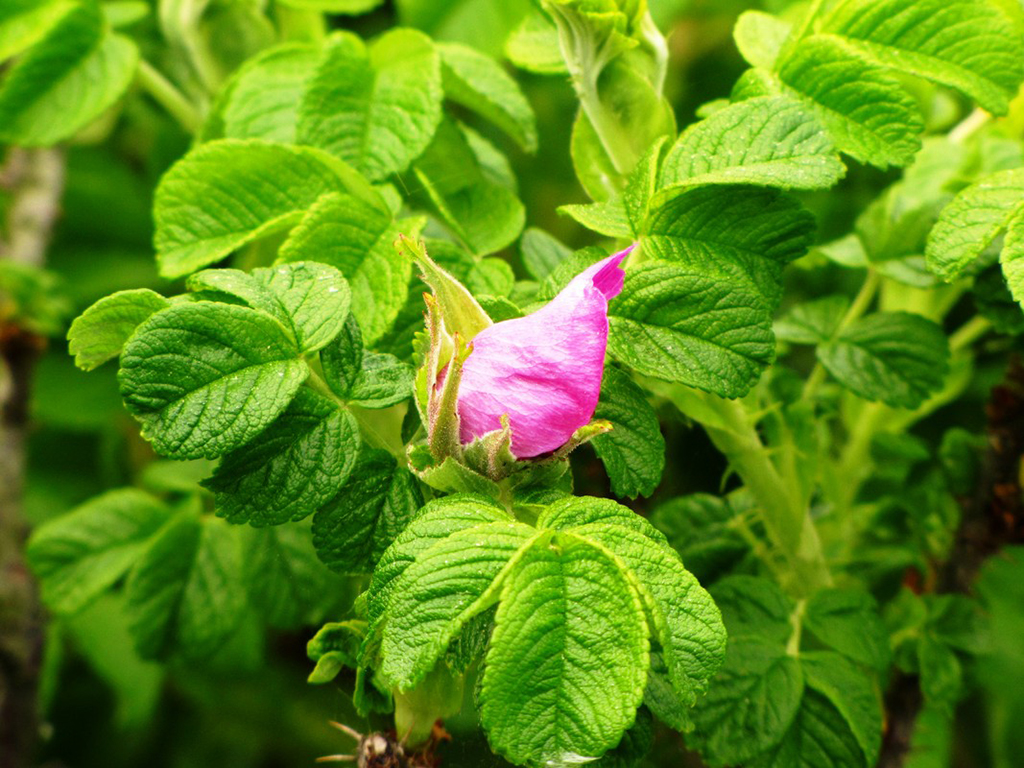}} &
\raisebox{-.5\height}{\includegraphics[width=1.8cm,height=1.8cm,keepaspectratio]{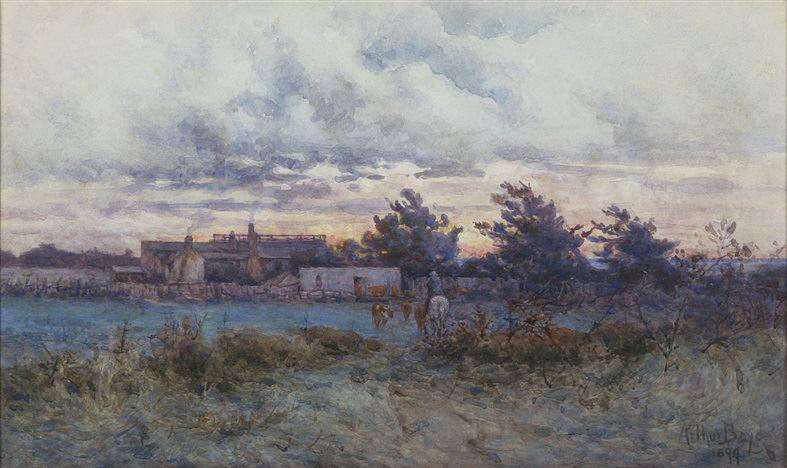}} \\
\midrule
{\footnotesize \raggedright \textbf{70th - 80th percentile}

2,648,181 items

21\% of PD12M} &
\raisebox{-.5\height}{\includegraphics[width=1.8cm,height=1.8cm,keepaspectratio]{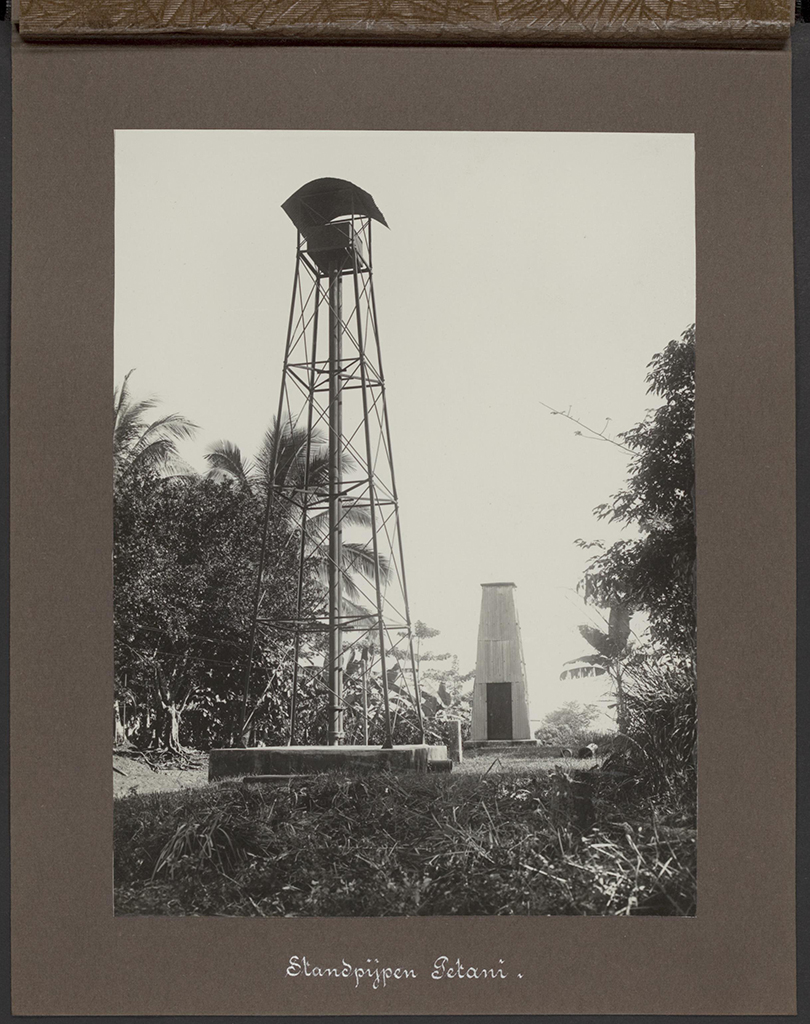}} &
\raisebox{-.5\height}{\includegraphics[width=1.8cm,height=1.8cm,keepaspectratio]{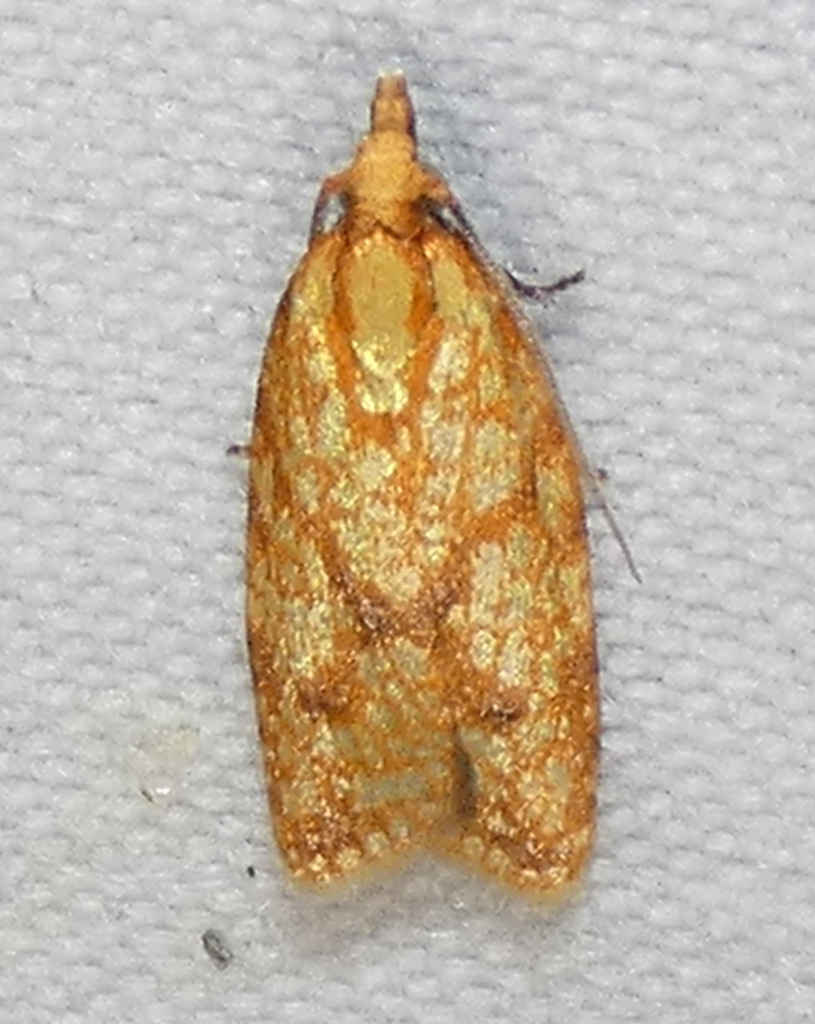}} &
\raisebox{-.5\height}{\includegraphics[width=1.8cm,height=1.8cm,keepaspectratio]{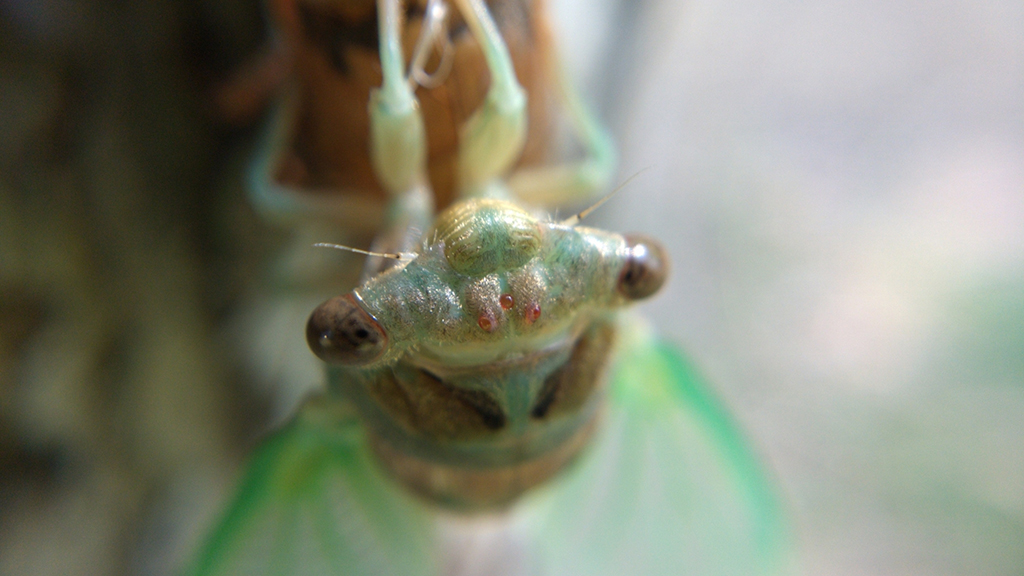}} &
\raisebox{-.5\height}{\includegraphics[width=1.8cm,height=1.8cm,keepaspectratio]{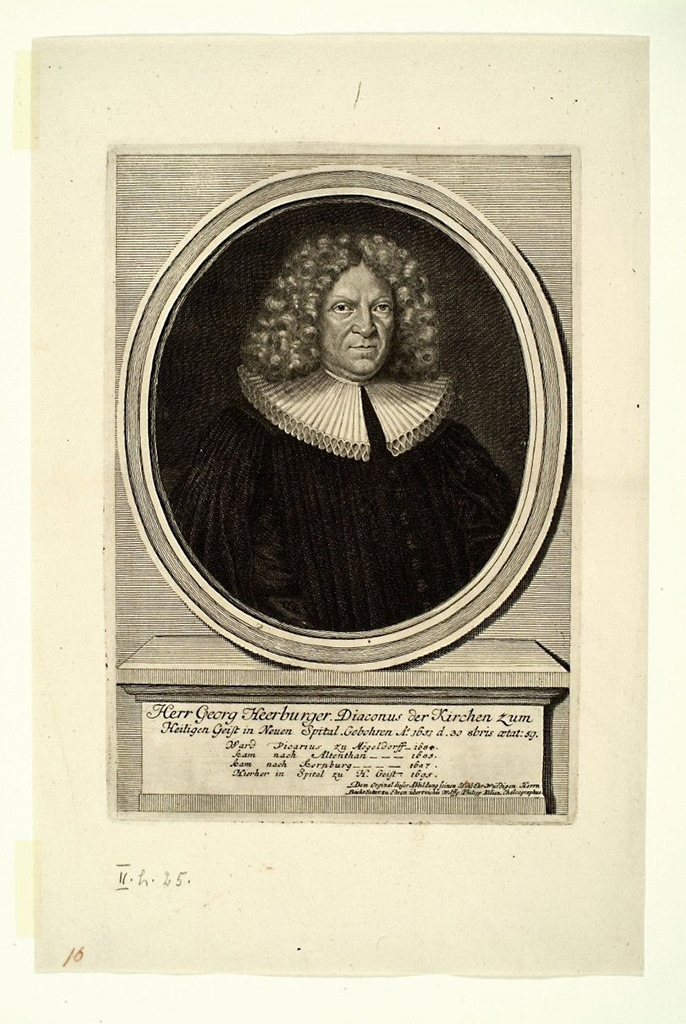}} &
\raisebox{-.5\height}{\includegraphics[width=1.8cm,height=1.8cm,keepaspectratio]{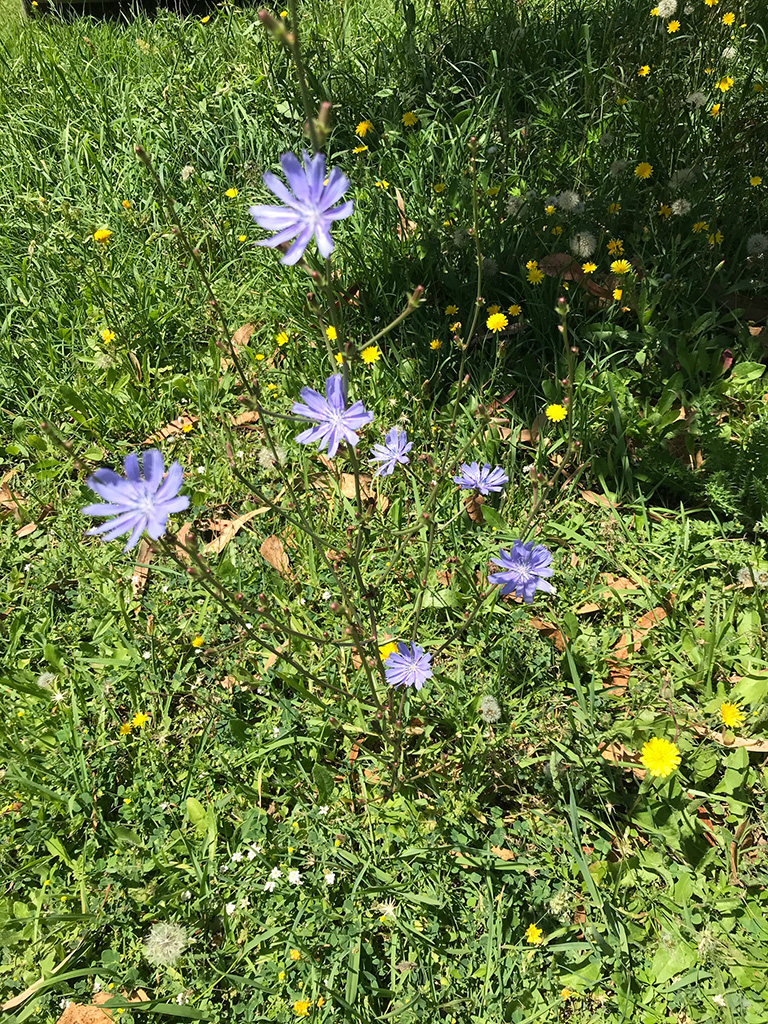}} &
\raisebox{-.5\height}{\includegraphics[width=1.8cm,height=1.8cm,keepaspectratio]{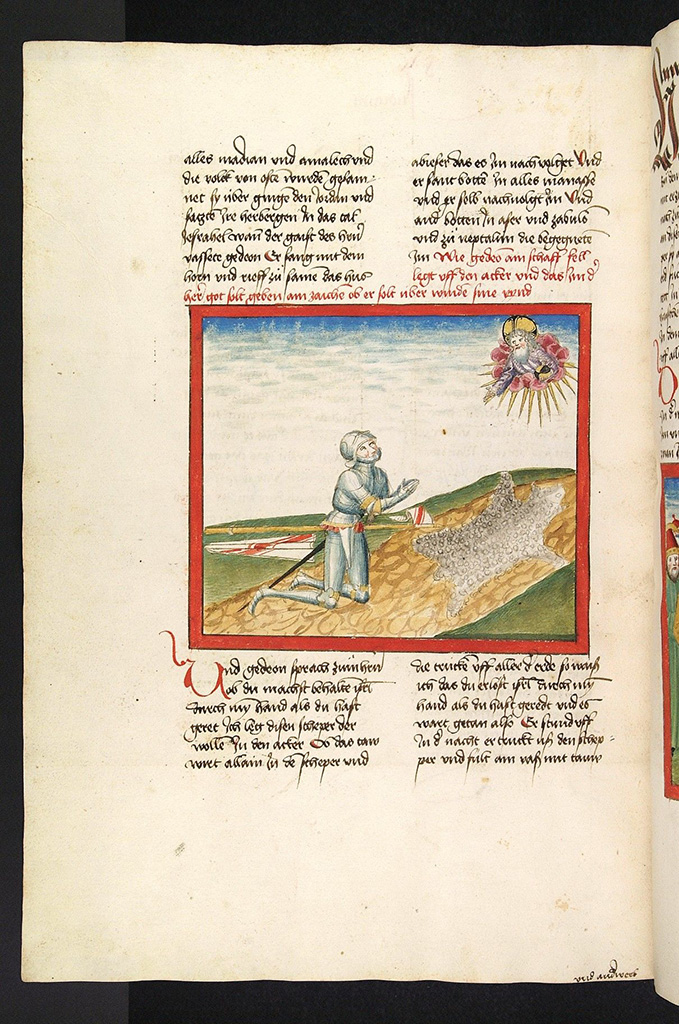}} \\
\midrule
{\footnotesize \raggedright \textbf{60th - 70th percentile}

2,078,873 items

17\% of PD12M} &
\raisebox{-.5\height}{\includegraphics[width=1.8cm,height=1.8cm,keepaspectratio]{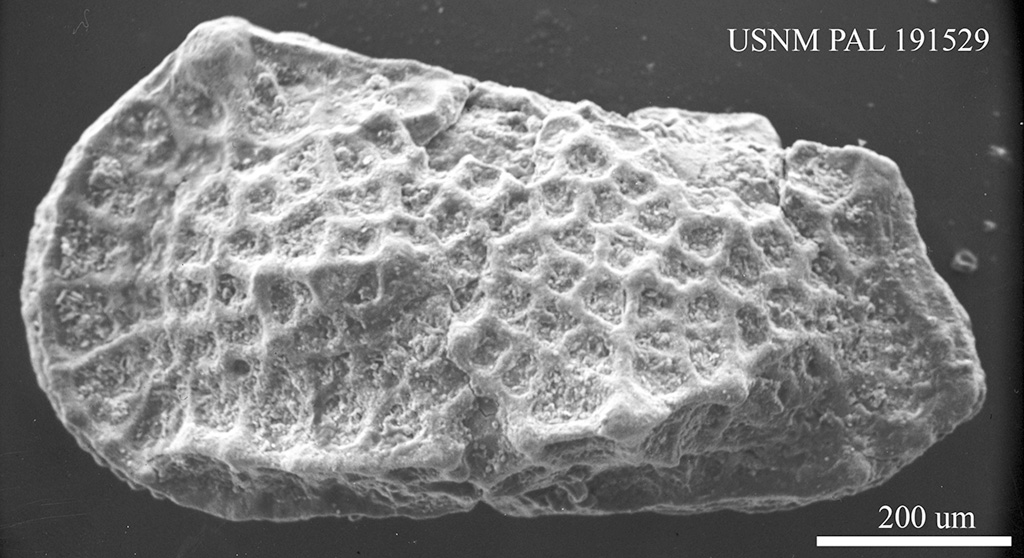}} &
\raisebox{-.5\height}{\includegraphics[width=1.8cm,height=1.8cm,keepaspectratio]{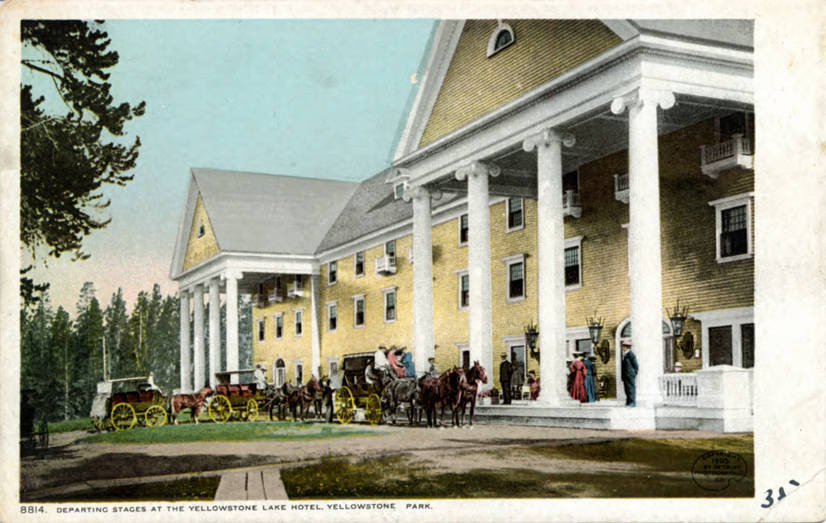}} &
\raisebox{-.5\height}{\includegraphics[width=1.8cm,height=1.8cm,keepaspectratio]{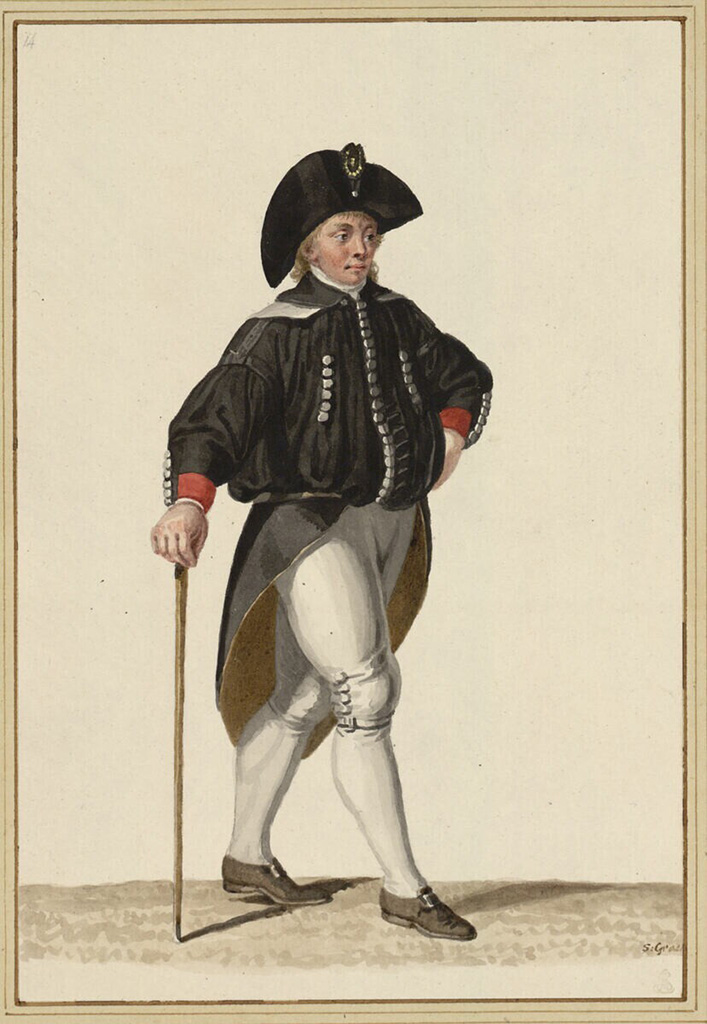}} &
\raisebox{-.5\height}{\includegraphics[width=1.8cm,height=1.8cm,keepaspectratio]{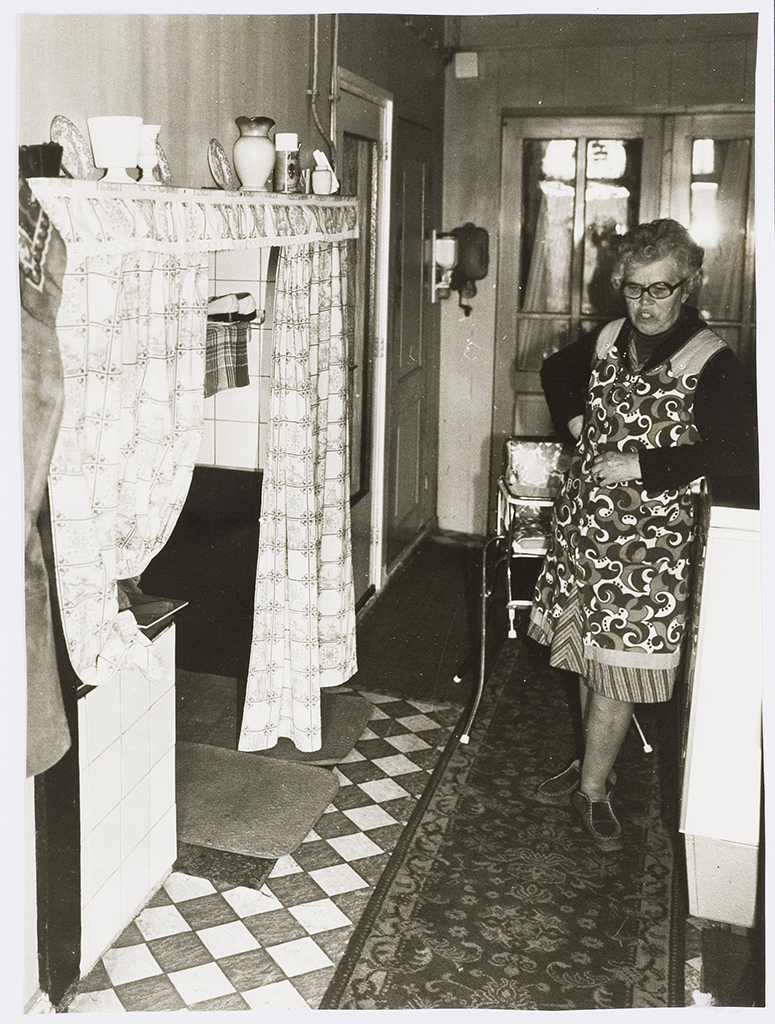}} &
\raisebox{-.5\height}{\includegraphics[width=1.8cm,height=1.8cm,keepaspectratio]{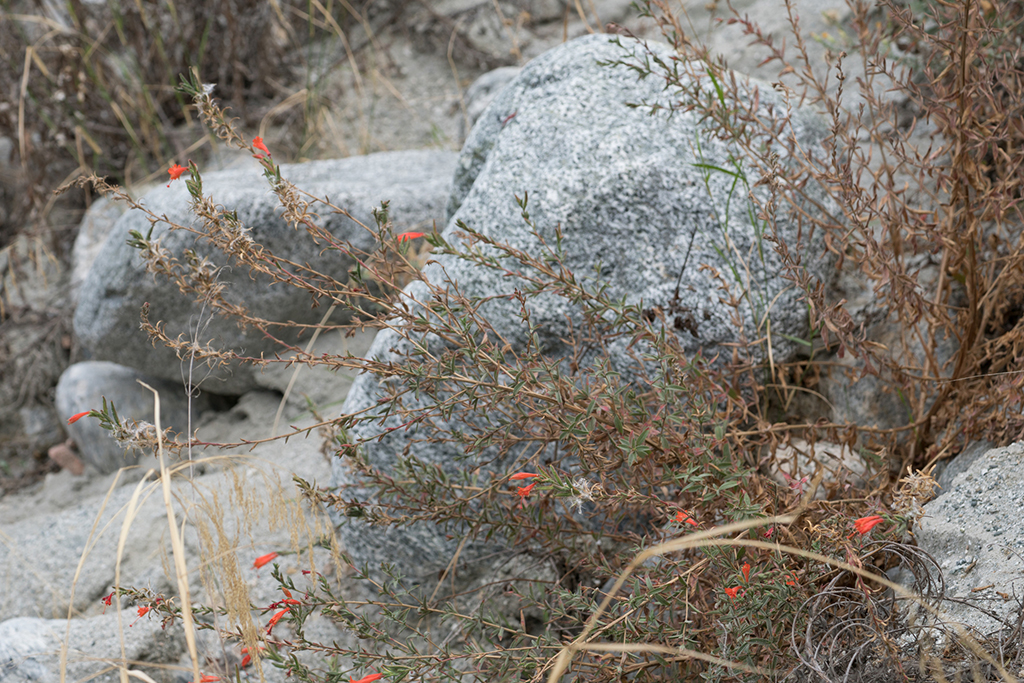}} &
\raisebox{-.5\height}{\includegraphics[width=1.8cm,height=1.8cm,keepaspectratio]{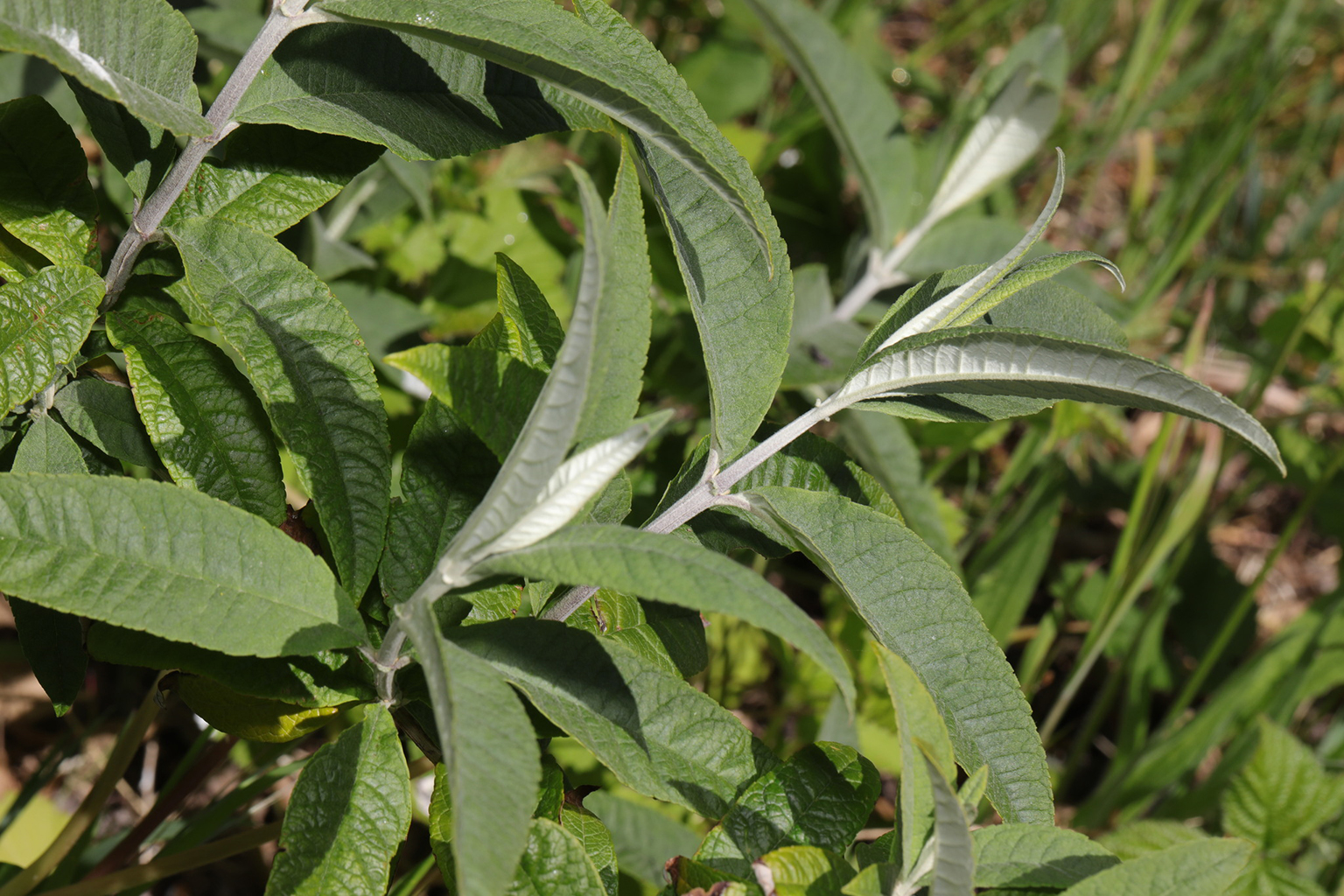}} \\
\midrule
{\footnotesize \raggedright \textbf{50th - 60th percentile}

1,310,279 items

10\% of PD12M} &
\raisebox{-.5\height}{\includegraphics[width=1.8cm,height=1.8cm,keepaspectratio]{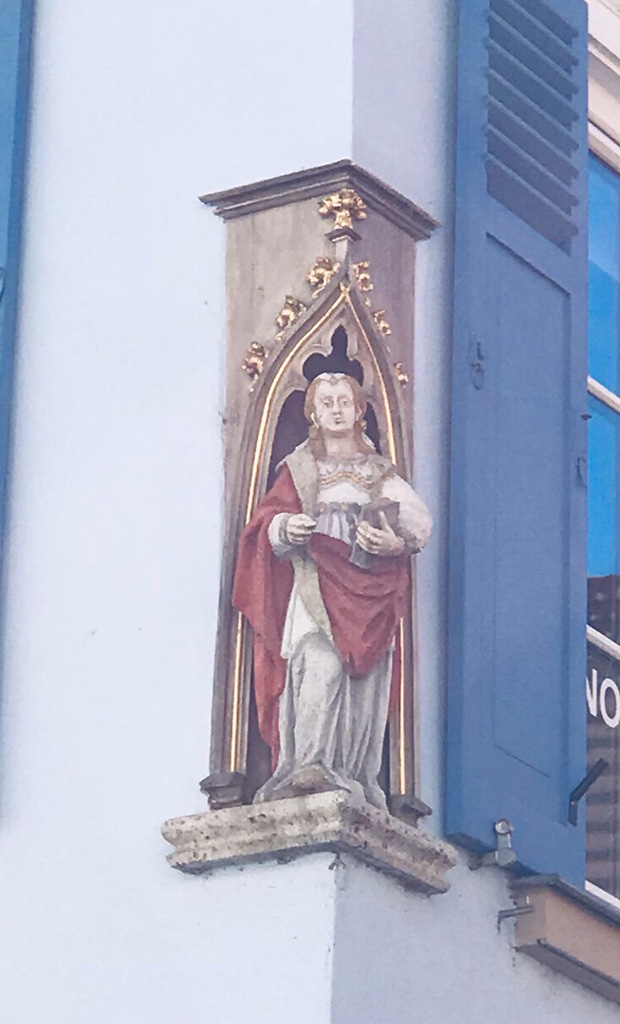}} &
\raisebox{-.5\height}{\includegraphics[width=1.8cm,height=1.8cm,keepaspectratio]{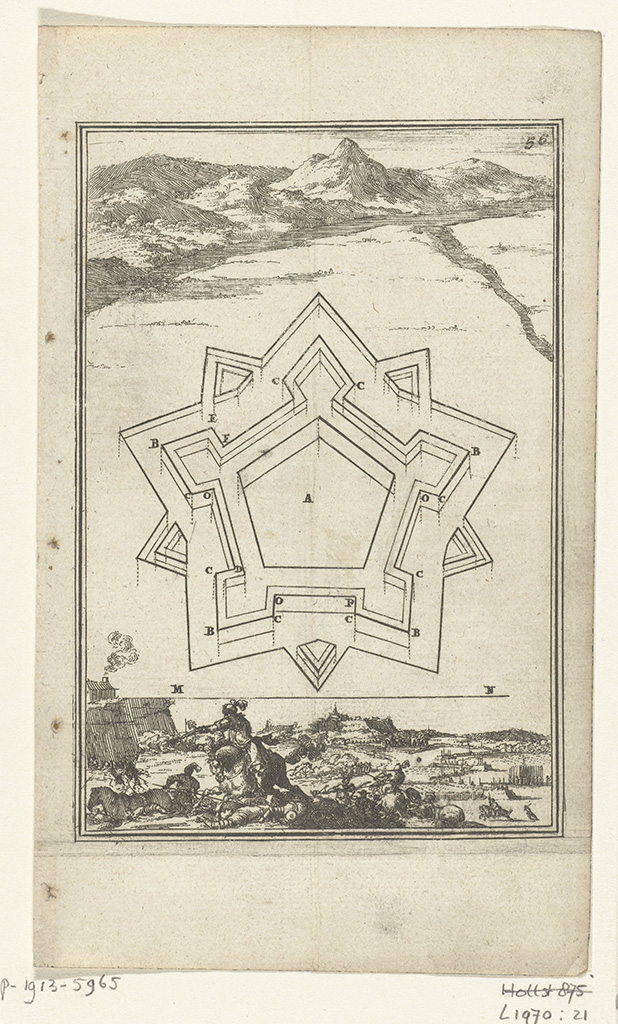}} &
\raisebox{-.5\height}{\includegraphics[width=1.8cm,height=1.8cm,keepaspectratio]{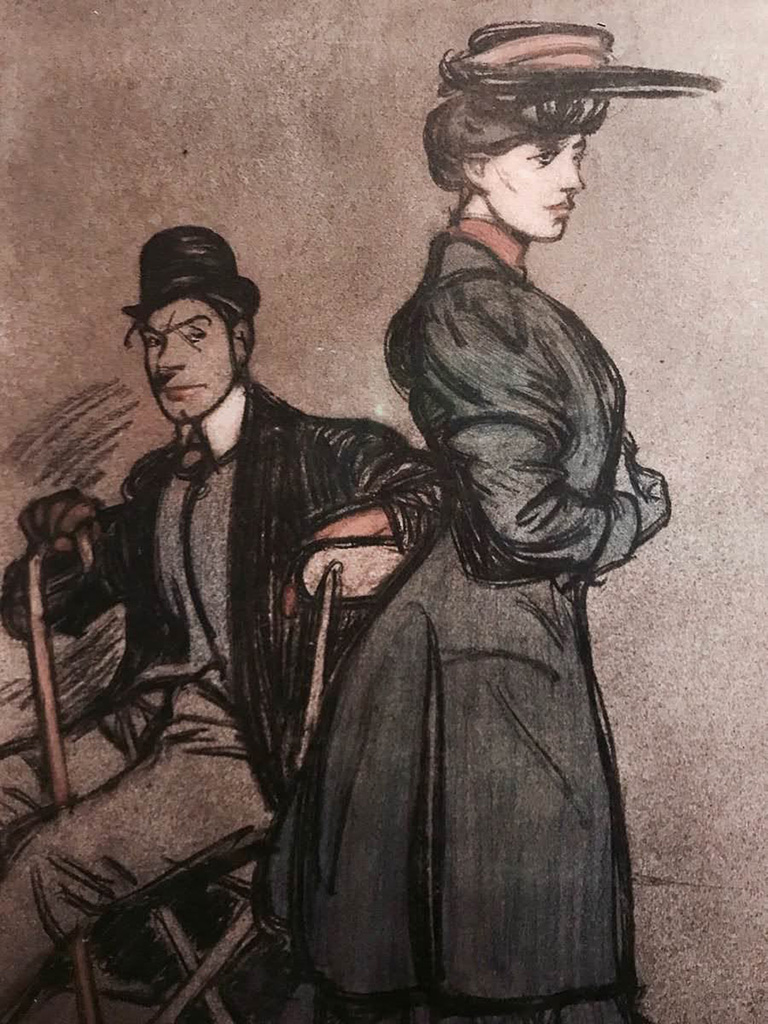}} &
\raisebox{-.5\height}{\includegraphics[width=1.8cm,height=1.8cm,keepaspectratio]{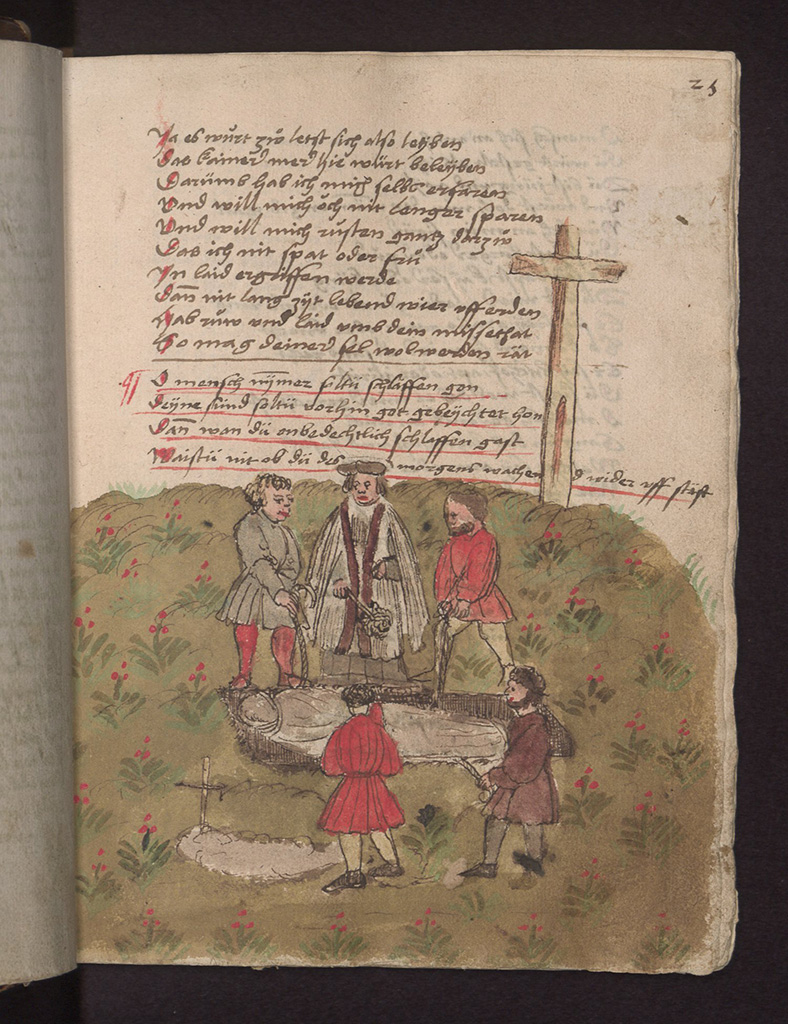}} &
\raisebox{-.5\height}{\includegraphics[width=1.8cm,height=1.8cm,keepaspectratio]{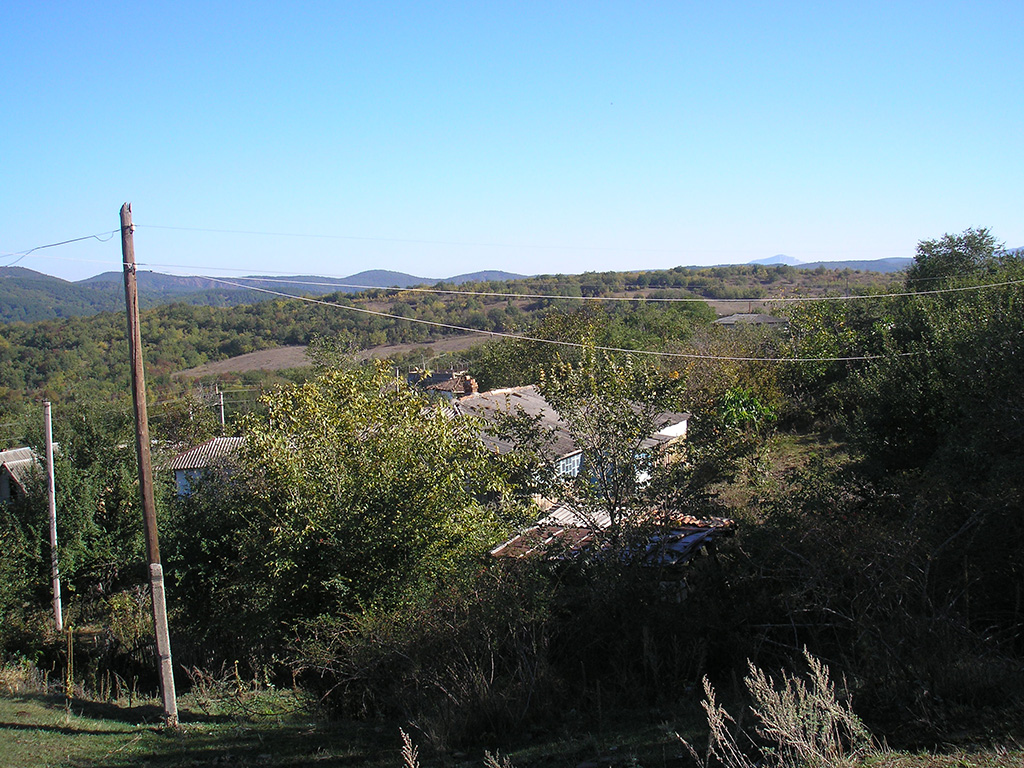}} &
\raisebox{-.5\height}{\includegraphics[width=1.8cm,height=1.8cm,keepaspectratio]{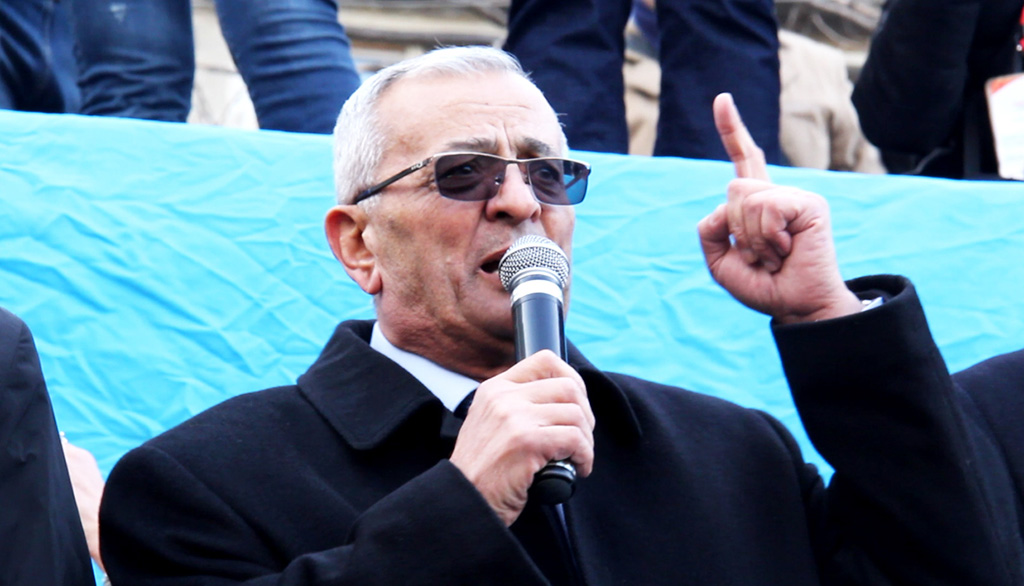}} \\
\midrule
{\footnotesize \raggedright \textbf{<50th percentile}

Excluded from PD12M} &
\raisebox{-.5\height}{\includegraphics[width=1.8cm,height=1.8cm,keepaspectratio]{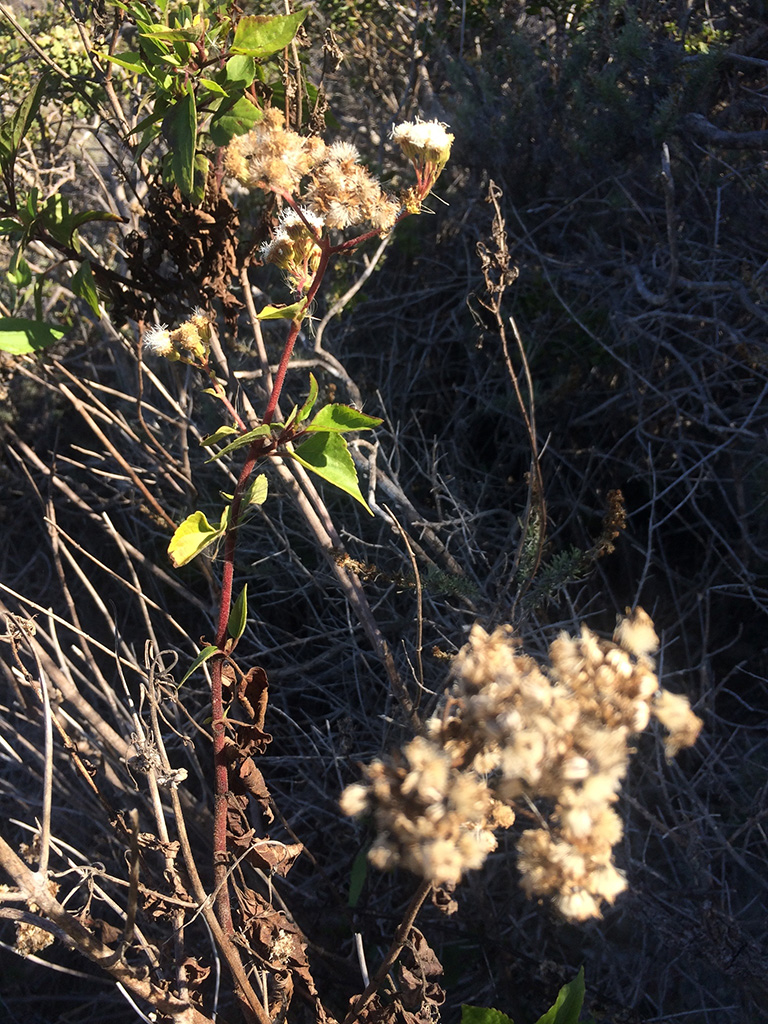}} &
\raisebox{-.5\height}{\includegraphics[width=1.8cm,height=1.8cm,keepaspectratio]{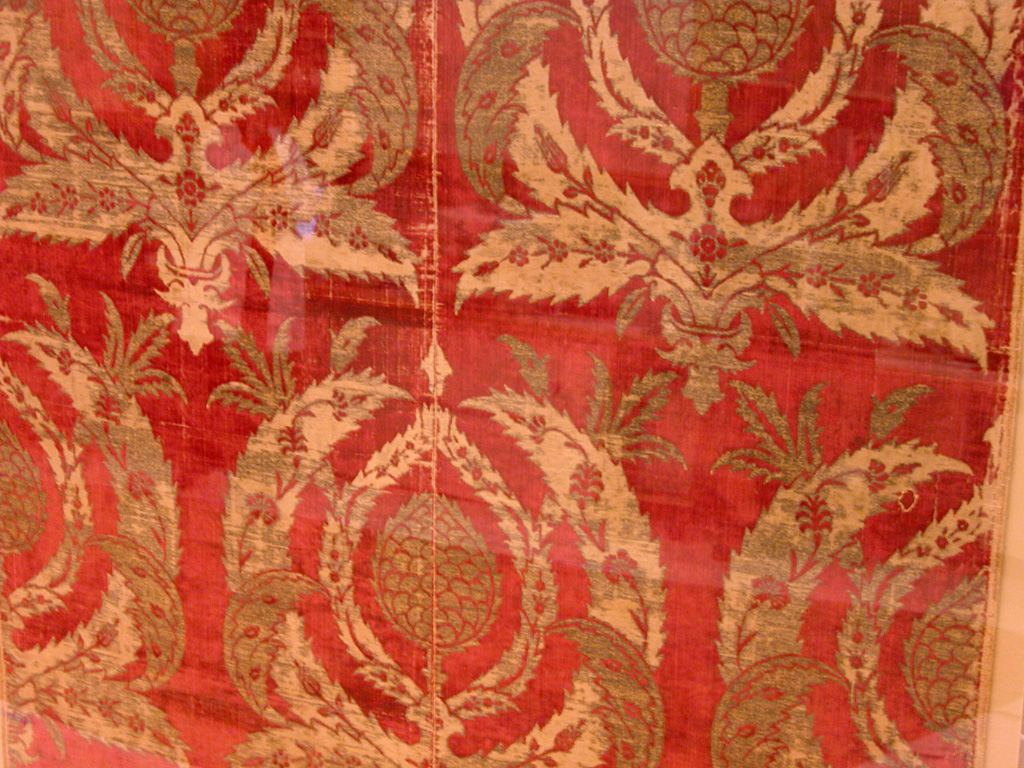}} &
\raisebox{-.5\height}{\includegraphics[width=1.8cm,height=1.8cm,keepaspectratio]{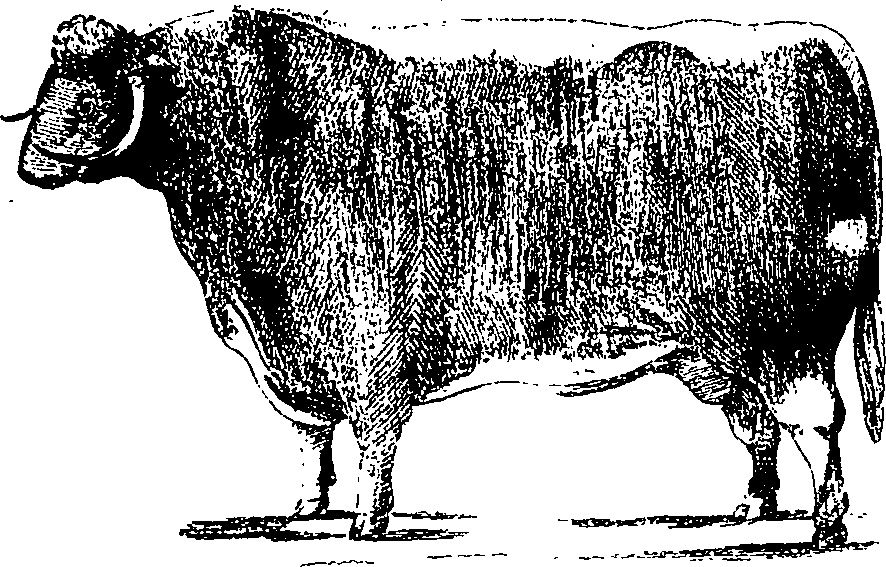}} &
\raisebox{-.5\height}{\includegraphics[width=1.8cm,height=1.8cm,keepaspectratio]{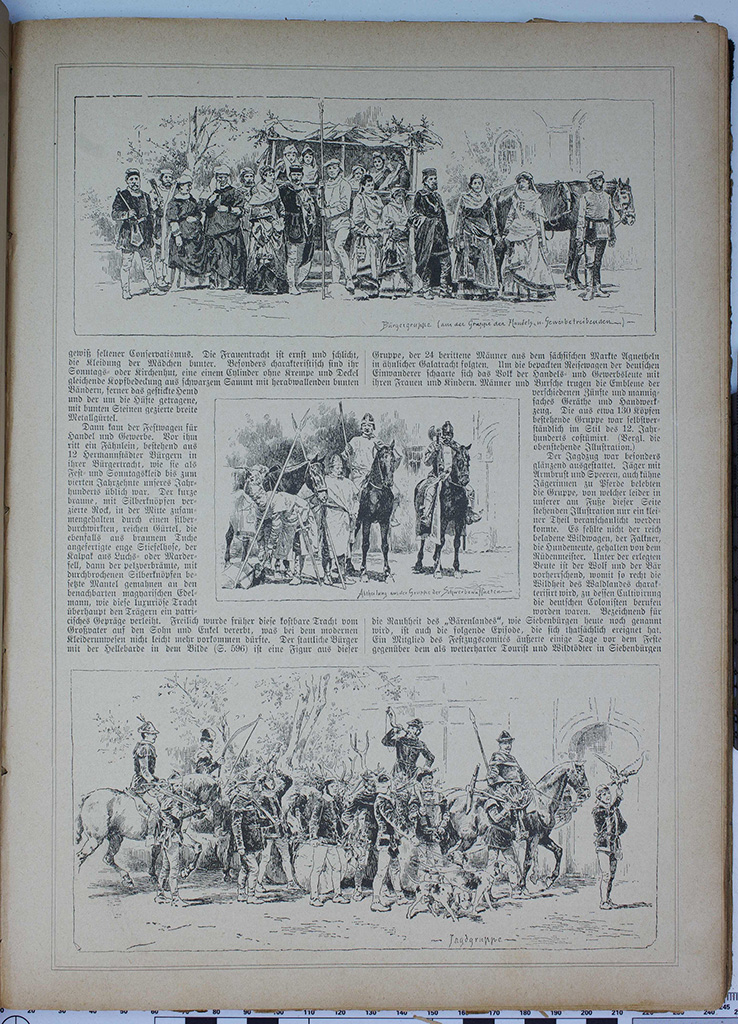}} &
\raisebox{-.5\height}{\includegraphics[width=1.8cm,height=1.8cm,keepaspectratio]{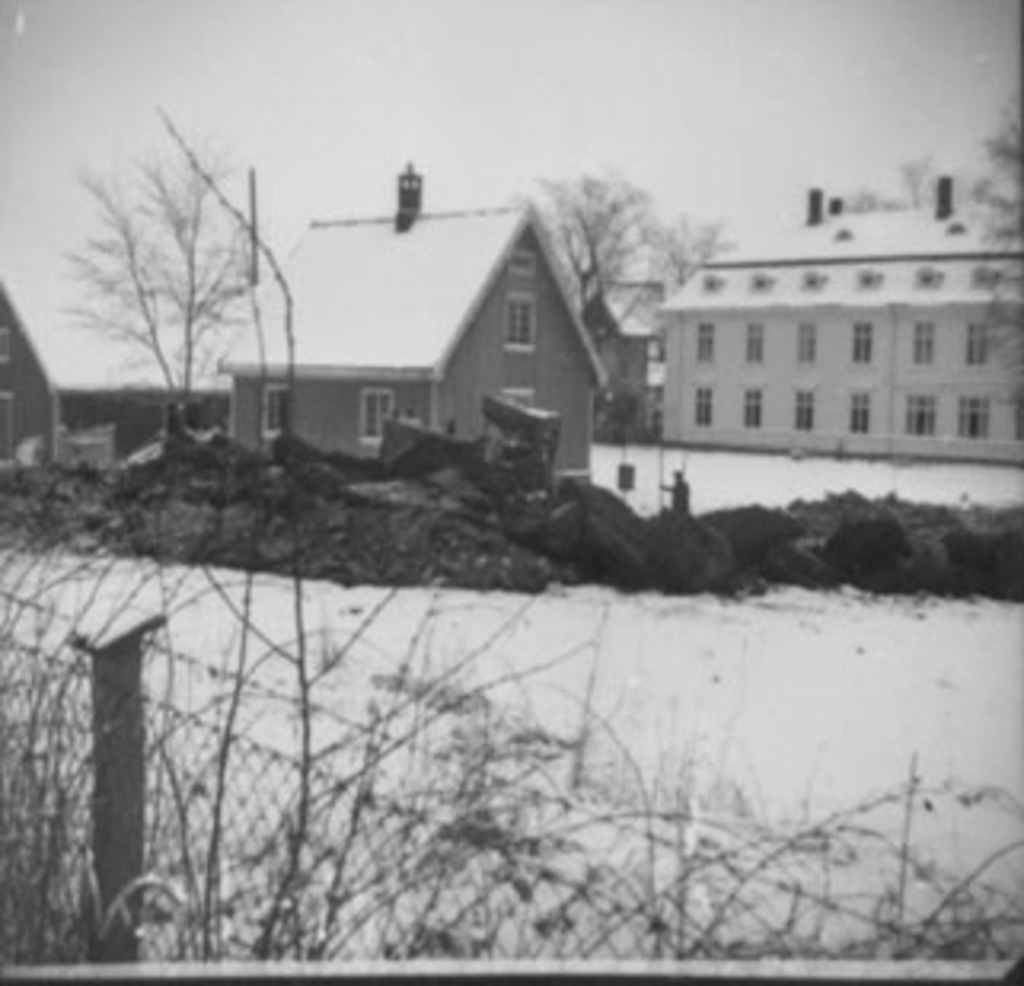}} &
\raisebox{-.5\height}{\includegraphics[width=1.8cm,height=1.8cm,keepaspectratio]{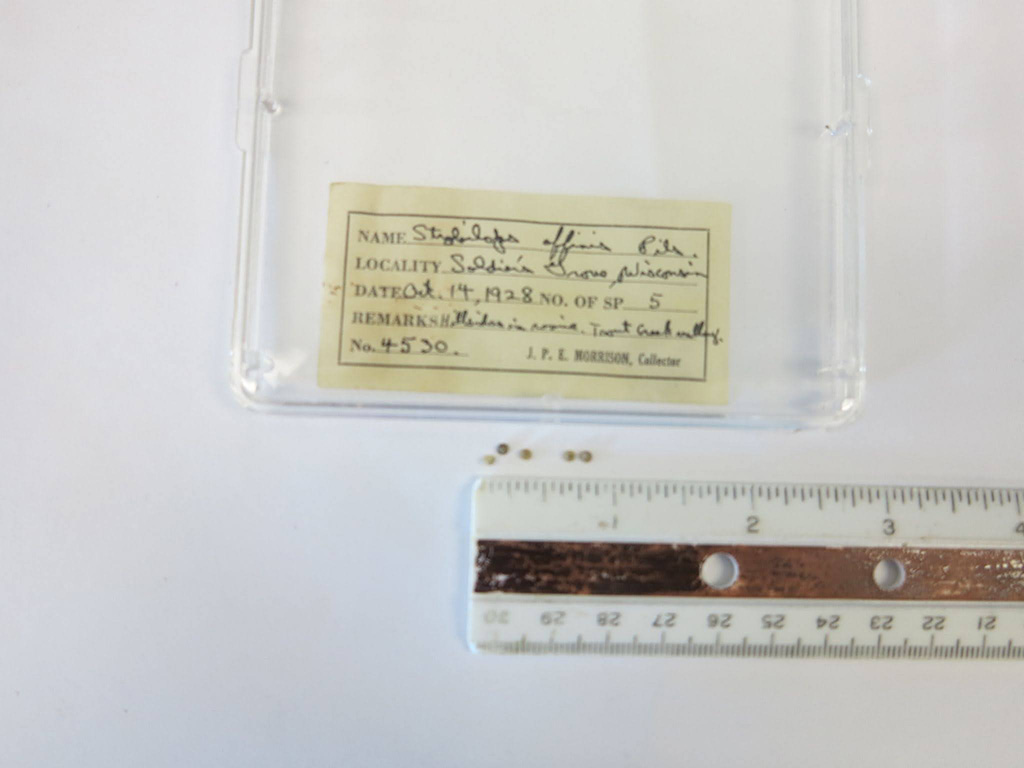}} \\
\bottomrule
\end{tabularx}
\label{tab:aesthetic_samples}
\end{table*}

\textbf{Manual Spot-checks:} Throughout the curation process, we performed spot-checks on random samples to verify the effectiveness of the automated filtering. We reviewed over 0.1\% of the dataset, with a focus on edge cases and copyright misclassifications. This process also led us to notice that nearly 2\% of Wikimedia Commons photographs had EXIF rotations \cite{geitgey2019exif}, which we corrected for our hosted copies. 

\subsection{Caption Generation}
 
We  generated detailed synthetic captions for each image using Florence-2-large \cite{xiao2023florence}. Captioning the $\sim$29M images not identified as document scans required 8 A100 80GB GPUs running for approximately 18 days.

We chose to use synthetic captions rather than the associated alt-text or image descriptions to maximize both privacy and quality. Large language models can memorize and reproduce PII from their training data \cite{carlini2019secret, carlini2021extracting}, although these cases are necessarily more rare than the rate of PII in the training datasets themselves. Similarly, synthetic captions might introduce "style leakage," whereby the captioner leaks details of a living artist's style by associating their name with public domain artwork that shares style characteristics. In cases of style leakage, a model could return consistent results when prompted with a living artist's name, giving the mistaken impression that the artist's work was included in the model's training data. We reduced the risk of style leakage by cross-referencing the synthetic captions against Midjourney's list of artists \cite{midjourney2023artists} to systematically remove any mention of artists still living after 1954.

\section{Data Governance Framework}
\label{sec:data_gov}

In addition to robust curation, dataset revisions are critical to remove problematic content that is discovered over time. Here, we outline three key mechanisms for dataset governance that enable datasets to continually improve: auditability, stabilization, and communication. We implement these mechanisms through Source.Plus, which provides technical infrastructure for exploring, reviewing, and refining PD12M while maintaining a detailed history of its changes. 

\subsection{Auditability}

Although PD12M follows strict copyright restrictions, the "Creative Commons fallacy" \cite{prabhu2020large} identifies that copyright status alone does not address privacy and safety issues. These issues are more likely to be discovered as more users explore the dataset, suggesting that dataset audits are valuable and most effective when they are public and community-driven \cite{jo2020lessons}. 
 
Dataset audits require supporting transparency, including making it possible to search and view the images used in the dataset regardless of its scale, and defining clear processes for flagging problematic items following a dataset's release.

Source.Plus provides keyword, semantic, faceted, and reverse image search capabilities to enable dataset exploration. Source.Plus also provides a public feedback mechanism through its "flag" feature. Users can flag items for any reason, including copyright, bias, and privacy concerns \cite{spawning2024takedown, spawning2024metadata}, and record their justification in a free-text field. When an item is flagged, it becomes immediately invisible to users pending review. Reviewed items are either restored to the dataset or replaced with a similar image. 

\subsection{Stabilization}

Training dataset stability is crucial for meaningful model comparisons \cite{pineau2021improving}. Even small changes in training data can significantly impact model performance \cite{quionero2009dataset}, and inconsistent datasets between research groups can disrupt reproducibility \cite{belz2021systematic}. Our approach ensures that necessary improvements can be made while preserving the statistical properties needed for reliable benchmarking and comparative analysis \cite{bouthillier2021accounting, rabanser2019failing}.

When an image requires replacement, we select an alternative from the $\sim$25 million additional public domain images in Source.Plus (as of October 2024) using quantitative measures, such as similarity between semantic and perceptual embeddings \cite{radford2021learning, zhang2018unreasonable}, aesthetic scores, image dimensions, file size, metadata completeness, and other key features. 

\subsubsection{Image Archiving}

Archiving the images under the control of the dataset maintainers improves stability, and is made possible because the images in PD12M are understood to be in the public domain. Open-web image links are susceptible to changes introduced by others \cite{carlini2023poisoning, zhang2024persistentpretrainingpoisoningllms} and would have to be regularly monitored to achieve comparable stability without image archiving.

Pew Research has shown that $\sim$38\% of webpages that existed in 2013 were no longer accessible a decade later \cite{pew2024online}. Recent changes in API policies, such as Flickr's restrictions on access to CC-licensed images \cite{flickr2023api}, and the growing trend of websites modifying their robots.txt files to limit crawling further hinder the consistency of previous datasets. The Data Provenance Initiative underscores this issue, reporting that between August 2023 and January 2024, data restriction requests increased dramatically, with up to 45\% of previously accessible content in major datasets becoming restricted \cite{longpre2023consent}. 

An often overlooked contributor to this trend is the increasing hosting costs associated with web-scraping  \cite{holscher2024crawlers, cox2023scraping}. Web-scraped datasets externalize egress costs to the original content hosts, often public institutions. Rising costs can discourage institutions from hosting data publicly, further limiting the future availability of resources for AI researchers and the public at large. Image archiving accounts for these costs and reduces the burden on public institutions.

\subsection{Communication}

Finally, effective use by the research community requires monitoring and publicizing changes to datasets over time. Each item in Source.Plus has its own changelog, allowing researchers to track the granular decisions that impact the dataset and allowing dataset maintainers to create digests to accompany new versions.

\section{Limitations and Future Work}

\textbf{Geographical, Cultural, and Historical Bias:} Digitized public domain content over-represents Western geographic regions and cultures \cite{torralba2011unbiased, mccarthy2024openglam, risam2019new}. It also over-represents older content, limiting the dataset's applicability to contemporary visual tasks \cite{crawford2021excavating}. 

Despite careful curation, offensive images reflecting these biases are very likely to persist in PD12M.  We will use the Source.Plus platform to address problematic contents as they are discovered, and we welcome feedback. We are also seeking sources of public domain media having a more diverse representation, and we encourage readers to contact us with suggestions. 

\textbf{PD12M Metadata:} In the initial release of PD12M, we include images and captions, which are sufficient for training text-to-image models. However, we do not include any of the additional metadata that we collected from the image sources. While that metadata is searchable on Source.Plus, we currently recommend against using it for commercial model training. Some source institutions provide images with a PD or CC0 designation alongside metadata elements that carry a more restrictive CC license. Future versions of PD12M will include additional metadata elements as we determine whether they share a public domain designation.

\textbf{Synthetic Captions:} We created the captions for PD12M using only images as prompts. As we release additional metadata, providing a captioning model with more context (e.g., artistic medium) will likely improve the specificity and accuracy of the captions.

We used Florence-2-large \cite{xiao2023florence}, which was trained using copyrighted materials with limited transparency about its training data. While we reviewed and edited the captions it produced to eliminate reference to non-public domain artists, we would ultimately like to replace all captions using a model trained entirely on public domain materials. At the time of writing, no such public domain-exclusive captioning models are known to the authors.

\section{Conclusion}

PD12M is a highly aesthetic, public domain alternative to existing image-text datasets. It addresses critical licensing concerns and demonstrates the feasibility of incorporating responsible AI recommendations at web-scale. Source.Plus supports ongoing dataset governance to address bias, privacy, and safety by allowing responsive data changes that don't compromise dataset reproducibility. 

With Source.Plus, we aim to promote responsible AI development practices across the industry \cite{bender2021dangers}. We hope these efforts offer an opportunity for individuals and organizations who are uneasy with current dataset practices to participate in AI development \cite{gilman2023democratizing, seger2023democratising}.

\section{Acknowledgments}

We express our sincere gratitude to the numerous institutions and individuals who have contributed to the public domain. Their commitment to open access and sharing of knowledge has made this work possible. A full list of contributing publishers and institutions can be found in Appendix \ref{app:summary}.

This research was made possible in part by generous support from Google Cloud Platform (GCP), which provided cloud computing credits that were utilized for the majority of the inference tasks in this project. The authors also gratefully acknowledge the AWS Open Data Sponsorship program for hosting the image files, making this large-scale dataset readily accessible to the research community.

\bibliography{main}

\begin{appendix}
\appendix
\section {Appendices}
\subsection{Dataset Summary}
\label{app:summary}
We maintain an up-to-date \href{https://docs.google.com/spreadsheets/d/18HmLi0QycFdVyZUnhrZ5EgqteyZhGiVs_-MdhVqT15M/edit?gid=0#gid=0}{summary of the PD12M dataset} contents via a public Google sheet. This summary includes counts by contributing institutions as well as enumerated content filters.

\subsection{Legal Disclaimer}
\label{app:disclaimer}
To the best of our knowledge and available resources, all works included in PD12M were labeled with a Public Domain mark or CC0 license. While we have implemented conservative filtering regarding copyright status, due to the scale of the dataset (12.4 million items), we cannot guarantee with absolute certainty the licensing status of any individual work.

If you identify any work that you believe has been incorrectly included in PD12M due to licensing status, please notify us through our formal \href{https://source.plus/takedown-policy}{Takedown Policy}. Upon review, any erroneously included work will be promptly removed and replaced following our dataset maintenance procedures.

For organizations seeking to build upon PD12M, we recommend conducting an independent verification of licensing status appropriate to use case and jurisdiction.

The PD12M dataset is released under the Community Data License Agreement - Permissive 2.0 (\href{https://cdla.dev/permissive-2-0/}{CDLA-Permissive-2.0}). Under this license, you are free to use, modify, and share this dataset, provided you comply with the terms of the CDLA-Permissive-2.0 license.

LEGAL NOTICE: This disclaimer and the information contained herein do not constitute legal advice. Any person or organization seeking to use this dataset should consult with their own legal counsel regarding their rights and obligations. The authors and maintainers of PD12M are not law firms and cannot provide legal advice.

THE DATASET IS PROVIDED "AS IS", WITHOUT WARRANTY OF ANY KIND, EXPRESS OR IMPLIED, INCLUDING BUT NOT LIMITED TO THE WARRANTIES OF MERCHANTABILITY, FITNESS FOR A PARTICULAR PURPOSE, AND NONINFRINGEMENT. IN NO EVENT SHALL THE AUTHORS OR COPYRIGHT HOLDERS BE LIABLE FOR ANY CLAIM, DAMAGES OR OTHER LIABILITY, WHETHER IN AN ACTION OF CONTRACT, TORT OR OTHERWISE, ARISING FROM, OUT OF OR IN CONNECTION WITH THE DATASET OR THE USE OR OTHER DEALINGS IN THE DATASET.
\end{appendix}

\end{document}